\documentclass[pdflatex,sn-mathphys-num]{sn-jnl}

\usepackage{graphicx}%
\usepackage{multirow}%
\usepackage{amsmath,amssymb,amsfonts}%
\usepackage{amsthm}%
\usepackage{mathrsfs}%
\usepackage[title]{appendix}%
\usepackage{textcomp}%
\usepackage{manyfoot}%
\usepackage{booktabs}%
\usepackage{algorithm}%
\usepackage{algorithmicx}%
\usepackage{algpseudocode}%
\usepackage{listings}%
\usepackage{overpic}

\usepackage{tabularx}
\usepackage[table]{xcolor}
\usepackage[T1]{fontenc}
\usepackage{newtxtext}
\usepackage{newtxmath}
\usepackage{fancyhdr}

\theoremstyle{thmstyleone}%
\theoremstyle{thmstyletwo}%

\theoremstyle{thmstylethree}%

\begin{document}

\title[Human-like-plans]{Understanding Human-like Solutions in Combinatorial Optimization via Learning and Search}


\author[1]{\fnm{Haijiang} \sur{Yan}}\email{haijiang.yan@warwick.ac.uk}
\equalcont{These authors contributed equally to this work.}

\author[2]{\fnm{Jian-Qiao} \sur{Zhu}}\email{zhujq@hku.hk}
\equalcont{These authors contributed equally to this work.}

\author[3]{\fnm{Liqiang} \sur{Huang}}\email{lqhuang@cuhk.edu.hk}

\author*[4,5]{\fnm{Ming} \sur{Meng}}\email{mengm@uab.edu}

\affil[1]{\orgdiv{Department of Psychology}, \orgname{The University of Warwick}, \state{Coventry}, \country{United Kingdom}} 

\affil[2]{\orgdiv{Department of Psychology}, \orgname{The University of Hong Kong}, \state{Hong Kong}, \country{China}} 

\affil[3]{\orgdiv{Department of Psychology}, \orgname{The Chinese University of Hong Kong}, \state{Hong Kong}, \country{China}}

\affil[4]{\orgdiv{School of Psychology}, \orgname{South China Normal University}, \state{Guangdong}, \country{China}}

\affil*[5]{\orgdiv{Department of Neurobiology}, \orgname{The University of Alabama at Birmingham}, \state{Alabama}, \country{United States}}



\abstract{Humans often find good solutions to combinatorial optimization problems that are computationally hard even for advanced computer algorithms. In the Euclidean traveling salesman problems (TSP), people rapidly produce tours that are near-optimal, despite severe limits on time and computation. What makes a tour human-like, and how might such solutions be learned? Here we address these questions through a large-scale behavioral and computational investigation of human performance in Euclidean TSP. We sampled a broad space of TSP instances, collected human solutions, and compared them with neural policies based on Pointer Networks, which are recurrent neural networks with an attention-based pointing mechanism that define probability distributions over valid tours. We trained these networks under multiple objectives, including reinforcement learning (RL), supervised learning from optimal tours, supervised learning from human tours, and RL fine-tuning after optimal-supervised pretraining. Human tours were not identical to optimal tours, but occupied a near-optimal geometric basin: they shared many structural properties with optimal solutions while preserving systematic human-specific deviations. The best account of human tours was not direct imitation of optimal tours, but a model pretrained on optimal tours, fine-tuned by RL, and decoded through $\text{Best-of-}N$ sampling. These findings suggest that human-like solutions may emerge from a combination of structured supervised learning, RL, and test-time search, echoing computational principles underlying many modern artificial intelligence systems.}




\maketitle

\section{Introduction}\label{sec1}

\begin{figure}[h!]
    \centering
    \begin{minipage}{0.45\linewidth}
        \begin{overpic}[width=\linewidth]{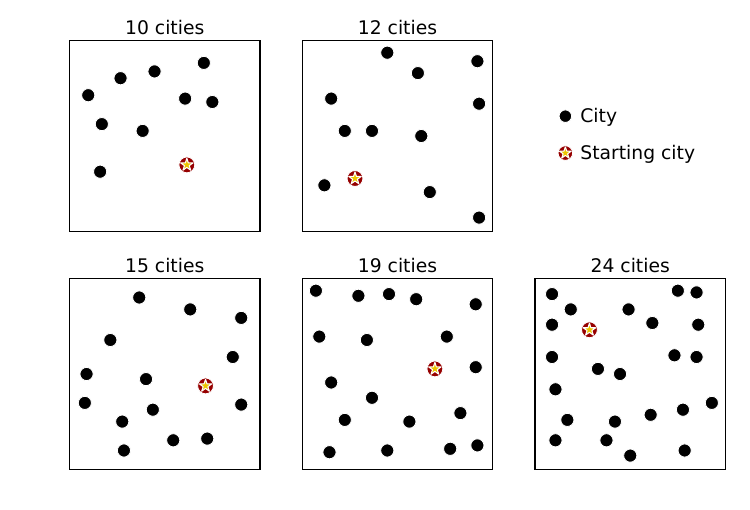}
            \put(-5,65){\Large\bfseries a}
        \end{overpic}
    \end{minipage}
    \begin{minipage}{0.40\linewidth}
        \begin{overpic}[width=\linewidth]{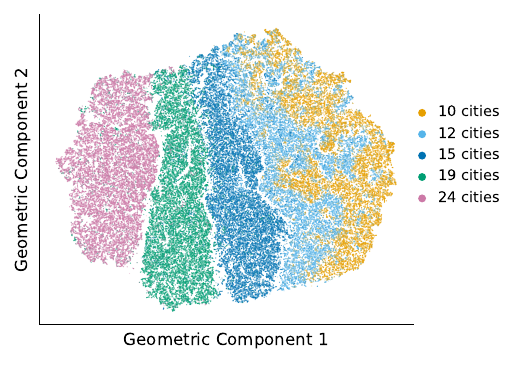}
            \put(-5,70){\Large\bfseries b}
        \end{overpic}
    \end{minipage}
    \caption{
    Euclidean TSP instances from the \texttt{tsp150k} dataset.
    \textbf{(a)} Example TSP instances from each of the five complexity levels. Star symbols denote the starting cities. In each trial, participants were instructed to begin at the designated starting city, visit every city exactly once, and return to the starting city to complete the tour. 
    \textbf{(b)} Visualization of the TSP instance space. Each instance in the dataset is represented by an eight-dimensional vector of geometric features (see Appendix \ref{ap:geo_properties}) extracted from the tours produced by human participants for that instance. Feature vectors were averaged across human tours for each instance. We then applied t-distributed stochastic neighbor embedding (t-SNE) to project the resulting feature vectors into a two-dimensional space for visualization. Each dot corresponds to a single TSP instance, with colors indicating its complexity level.
    }
    \label{fig:showcase_patterns}
\end{figure}

Many combinatorial optimization problems are deceptively simple: easy to state, but hard to solve. This challenge is not confined to mathematics and computer science. Many everyday cognitive tasks also have a combinatorial structure. Planning errands, cooking a meal, packing a suitcase, or organizing a workday all require selecting and ordering actions from a large space of possibilities. The Euclidean Traveling Salesman Problem (TSP) is a canonical example. The task is to find the shortest tour, measured by Euclidean length, that visits each city in a two-dimensional space exactly once and returns to the starting city (see example TSP instances in Figure \ref{fig:showcase_patterns}a). Although this goal is easy to understand, the problem is difficult because the number of possible tours grows factorially with the number of cities, making exhaustive search rapidly infeasible. Formally, the TSP is NP-hard, meaning that no polynomial-time algorithm is currently known that can guarantee optimal solutions for all instances. The TSP therefore provides a useful task for studying how intelligent agents produce good solutions under severe computational constraints.

What makes the TSP especially intriguing is that humans often produce near-optimal tours within only a few seconds \cite{macgregor2011human, macgregor1996human, van2003convex, kyritsis2018human}. This striking ability raises a central question for cognitive science and artificial intelligence: how do humans construct such high-quality solutions under severe computational constraints? More specifically, what cognitive algorithm supports near-optimal human TSP performance? Can this algorithm be approximated or implemented in computational models? And what geometric properties distinguish human-like tours from other possible solutions?

To better understand human-like solutions to the TSP, we conducted the largest behavioral experiment on human TSP performance to date and generated the \texttt{tsp150k} dataset, which spans 150,000 unique TSP instances and includes over 20 million human-generated tours from 1,107 participants (see Appendix \ref{ap:experimental_method_participants} for details). For each TSP instance, between 92 and 207 human tours were collected, with an average of 139 human tours per instance. The TSP instances can be further categorized into five distinct complexity levels based on the number of cities tested: 10 (simplest), 12, 15, 19, and 24 (most complex, see illustrations in Figure \ref{fig:showcase_patterns}). 

This large-scale behavioral dataset of human TSP performance thus offers a unique opportunity to systematically characterize human behavior and investigate a broad class of computational models of the TSP, including heuristic search methods and neural-network-based learning methods. Briefly, our descriptive analyses of human TSP behavior replicate many classical findings in the cognitive psychology literature on the TSP, including that human tours are near-optimal, approximately 10\% longer than the optimal tour \cite{macgregor1996human, macgregor1999spatial}; and that tour optimality gradually decreases while time spent on the task increases with TSP complexity \cite{dry2006human, macgregor2011human}. We also uncover several novel behavioral patterns, including that, after ignoring previously visited cities, people’s next-city choices stabilize at a high level of local optimality; and that people spend substantially more time deliberating or planning at the initial position (see Appendix \ref{ap:stylized_facts} for details). Collectively, these results suggest that people may employ a general think-then-implement strategy, which enables them to achieve near-optimal performance within relatively short timeframes.



\section{Developing Models of Human-like Tours}

To further understand the cognitive mechanisms underlying human tours, we explored a broad space of computational models of the TSP. The scale of our \texttt{tsp150k} dataset enabled us to examine this large hypothesis space of candidate cognitive models, which can be broadly classified into two categories: (i) search-only methods and (ii) hybrid methods that combine search and learning (see Table \ref{tab:overview_all_models}).

\begin{table}[h!]
    \centering
    \caption{Overview of computational models of TSP tours.}
    \begin{tabularx}{\linewidth}{lXX}
        \hline
        Model & Learning & Search  \\ 
        \hline
        Concorde (\textit{optimal}) \cite{david2006concorde} & None & Exact optimization \\
        Nearest Neighbor \cite{golden1980approximate} & None & Local heuristic \\
        Convex Hull Cheapest Insertion \cite{golden1980approximate} & None & Geometric heuristic \\
        Largest Interior Angle \cite{norback1977geometric} & None & Geometric heuristic \\
        Elastic Net \cite{durbin1987analogue} & None & Gradient descent  \\
        \hline
        PointerNet policy: $p_\text{optimal}(\pi|\mathbf{x})$ &  Supervised learning on optimal tours & Greedy decoding  \\
        & & Beam search \\
        & & Best-of-$N$ sampling \\
        PointerNet policy: $p_\text{RL}(\pi|\mathbf{x})$ &  Reinforcement learning with negative Euclidean length as the reward & Greedy decoding  \\
        & & Beam search \\
        & & Best-of-$N$ sampling \\
        PointerNet policy: $p_\text{optimal+RL}(\pi|\mathbf{x})$ &  Supervised learning on optimal tours followed by reinforcement learning fine-tuning & Greedy decoding  \\
        & & Beam search \\
        & & Best-of-$N$ sampling \\
        \hline
        PointerNet policy: $p_\text{human}(\pi|\mathbf{x})$ &  Supervised learning on human tours & Greedy decoding  \\
        & & Beam search \\
        & & Best-of-$N$ sampling \\
        \hline
    \end{tabularx}
    \vspace{1mm} 
    \textit{Note.}  $\pi$ denotes a tour, and $\mathbf{x}$ denotes the sequence of cities in a TSP instance.
    \label{tab:overview_all_models}
\end{table}

Search is a fundamental component of solving combinatorial optimization problems. Its appeal in the Euclidean TSP is intuitive: the quality of a candidate tour can be evaluated efficiently by its Euclidean length, allowing candidate solutions to be readily ranked. The difficulty therefore lies not in evaluating candidate tours, but in generating high-quality ones, as the space of possible tours is too large to search exhaustively. In light of these computational challenges, numerous heuristic search methods have been proposed in operations research and computer science as alternatives to exhaustive search \cite{norback1977geometric, golden1980approximate, durbin1987analogue}. Nearest Neighbor is one of the simplest search heuristics: given a partial tour, it iteratively extends the tour by selecting the nearest unvisited city at each step \cite{golden1980approximate}. It is, however, a myopic algorithm that overemphasizes local optimality, often leading to inefficient and suboptimal tours. Other heuristic algorithms exploit global geometric structure to generate tours that more closely resemble optimal solutions. A prominent example is the Convex Hull heuristic, which first constructs an initial partial tour by connecting the cities on the boundary of the convex hull. The remaining interior cities are then inserted iteratively by selecting the insertion that minimizes the increase in tour length \cite{golden1980approximate} or maximizes the resulting interior angle \cite{norback1977geometric}, until all cities have been incorporated into the tour.

More sophisticated search algorithms approach the TSP by optimizing geometric or mathematical relaxations of the original combinatorial problem. One representative example is the Elastic Net \cite{durbin1987analogue}. It initializes a deformable ring consisting of more points than the number of cities at the center of the two-dimensional space. Each city attracts nearby points on the ring, while neighboring points exert elastic forces on one another to preserve the continuity of the ring. Together, these competing forces define an energy landscape, which is optimized using gradient descent to produce a locally optimal ring configuration. Once the optimization converges, the resulting ring is converted into a valid TSP tour. A related but exact approach is implemented in Concorde \cite{david2006concorde}. Concorde solves the TSP using a branch-and-cut algorithm that repeatedly optimizes a simplified mathematical formulation of the problem, progressively rules out invalid solutions, and systematically narrows the remaining search space until the globally optimal tour is certified. Concorde has solved TSP instances containing up to 85,900 cities \cite{david2006concorde} and was therefore used in this work to generate the optimal solutions for all instances in our dataset.

\begin{figure}[t!]
    \centering
    \includegraphics[width=\linewidth]{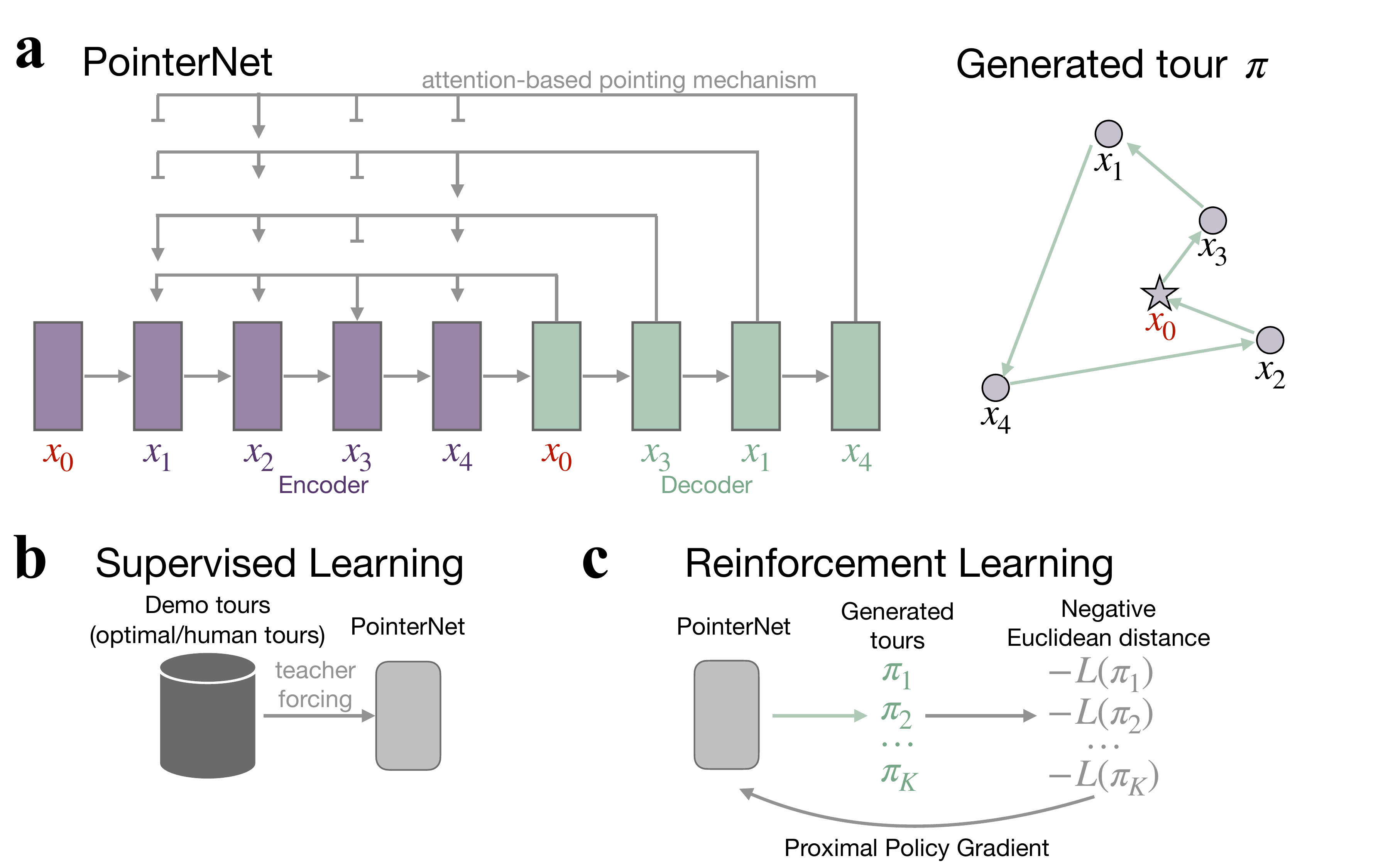}
    \caption{Illustration of the PointerNet architecture and training methods.
    \textbf{(a)} The PointerNet consists of two RNN modules. The encoder RNN (purple) sequentially processes the city locations, while the decoder RNN (green) autoregressively generates a city sequence. At each decoding step, an attention-based mechanism is used to select the next city while masking out previously decoded cities, thereby ensuring that the generated sequence forms a valid TSP tour. 
    \textbf{(b)} The PointerNet can be trained with supervised learning using teacher forcing on a set of demonstration tours \cite{williams1989learning}.
    \textbf{(c)} The PointerNet can also generate a set of candidate tours and receive reinforcement-learning signals that reward tours with shorter Euclidean lengths. The neural network is optimized using proximal policy optimization \cite{schulman2017proximal}.
    }
    \label{fig:pointer_net_illustration}
\end{figure}

An independent axis of improvement over search-only TSP solvers, and a model class that has been largely neglected in the psychological studies of TSP \cite{macgregor1996human, macgregor2011human, van2003convex}, is to \textit{learn} a proposal distribution that efficiently generates high-quality candidate tours for a given instance, thereby reducing the computational cost associated with search. This idea underlies modern neural-network-based TSP solvers, in which neural networks are trained to generate high-quality tours \cite{vinyals2015pointer, bello2016neural}. Here, we consider the Pointer Network (PointerNet hereafter) \cite{vinyals2015pointer}, a recurrent neural network (RNN) with an attention-based pointing mechanism that generates a valid permutation of the input sequence (see Figure \ref{fig:pointer_net_illustration} for an illustration and Appendix \ref{ap:computational_models} for details). The PointerNet first encodes the sequence of city coordinates into latent representations and then autoregressively decodes a tour. At each decoding step, the pointing mechanism selects one previously unvisited city, ensuring that every city is visited exactly once and that the decoded sequence constitutes a valid TSP tour. In our implementations, the starting city is always used as both the first encoded city and the first decoded city, mirroring our experimental design.

For PointerNet models of human TSP solutions, we can independently manipulate learning and search (see Table \ref{tab:overview_all_models}). Let $\mathbf{x}$ denote a city layout, and let $\pi$ denote a permutation of the cities representing the serial order of a tour. Different versions of the PointerNet were trained using different objectives, including (i) supervised learning on optimal tours, $p_\text{optimal}(\pi|\mathbf{x})$, (ii) RL with negative Euclidean tour length as the reward, $p_\text{RL}(\pi|\mathbf{x})$, and (iii) a combination of the first two: supervised learning on optimal tours followed by RL fine-tuning, $p_\text{optimal+RL}(\pi|\mathbf{x})$. We also trained a PointerNet directly on human tours, $p_\text{human}(\pi|\mathbf{x})$, which serves as an upper bound on model performance in predicting human behavior. As a result, each trained PointerNet defines a proposal distribution $p_\theta(\pi|\mathbf{x})$ (i.e., a stochastic policy of TSP), with different values of parameter $\theta$ corresponding to different training regimes. To decode a TSP tour from a trained model, we examined three search/decoding strategies: (i) greedy decoding, which myopically selects the most probable next city conditioned on the current partial tour; (ii) Beam search, which heuristically approximates the most probable complete sequence; and (iii) Best-of-$N$ sampling, which randomly samples $N$ complete tours and selects the shortest one among them. Overall, this results in a total of $4 ~(\text{learning objectives})\times 3 ~(\text{search strategies})=12$ PointerNet models of human TSP tours (see Appendix \ref{ap:computational_models} for details).

\section{Comparing Model-Generated with Human Tours}

All computational models considered here eventually generate valid TSP tours, with each model assuming a different underlying mechanism. We therefore developed a set of evaluation metrics to quantify the alignment between model-generated and human tours, each capturing a distinct aspect of model performance. As a human-level reference, we additionally included a human-tour baseline that randomly sampled one human solution for each TSP instance from the behavioral dataset. This reference incorporates inter-participant variability, a model that outperforms it should therefore capture systematic and coherent patterns of human behavior shared across individuals. Additionally, because the learning-based models (i.e., the PointerNet variants) were exposed to a training set, our model comparisons were conducted on the same held-out test set of TSP instances that were unseen during training (50,000 instances in total, with 10,000 at each complexity level).

The tour-level evaluation metrics aim to quantify the alignment between pairs of complete tours for a given TSP instance: one generated by a model and one produced by a human participant. We developed five such metrics of pairwise similarity: (i) Edge Overlap Ratio (EOR), which measures the proportion of shared edges between two tours, irrespective of their traversal order (e.g., the tours 1-2-3-4-5 and 1-4-5-2-3 share the edge 2-3 and 4-5); (ii) Longest Common Substring (LCS), which measures the length of the longest shared consecutive sequence of visited cities, regardless of where that sequence occurs within either tour (e.g., the tours 1-2-3-4-5-6-7 and 1-6-7-2-3-4-5 share the substring 2-3-4-5); (iii) Levenshtein distance (LD), which measures the minimum number of edits required to transform one tour into the other, providing a measure of overall sequence dissimilarity (e.g., the tours 1-2-3-4-5 and 1-2-3-5-4 differ only by substituting the last two positions, resulting in a small edit distance); (iv) Fréchet distance (FD), which measures the greatest spatial separation between corresponding points as the two tours are traversed from start to finish (e,g., two tours that share similar spatial trajectories but different visiting orders remain geometrically similar); and (v) Kendall's $\tau$, which measures the rank correlation between the visiting orders of cities in two tours (see Appendix \ref{ap:model_comparisons_evaluation_metrics} for details).

\begin{table}[t!]
    \centering
    \caption{Multi-dimensional evaluations between model-generated tours and human tours. All numbers were computed based on the same held-out test set.}
    \begin{tabularx}{\linewidth}{lXXXXXX}
         \hline
         Model & EOR~$\uparrow$  & LCS~$\uparrow$  & LD~$\downarrow$   & FD~$\downarrow$  & $\tau \uparrow$ & Aggregate score~$\uparrow$  \\
         \hline
         Concorde & \textbf{0.580}  & 0.488  & 6.156  & 39.442  & 0.331  & 0.592  \\
         Elastic Net & 0.572  & 0.484  & 5.940  & 38.688  & 0.332  & 0.593  \\
         Nearest Neighbor & 0.436  & 0.354  & 9.190  & 61.940  & 0.266  & 0.447  \\
         Convex Hull Cheapest Insertion & 0.571  & 0.418  & 10.437  & 65.257  & 0.130  & 0.449  \\
         Largest Interior Angle & 0.561  & 0.412  & 10.460  & 65.240  & 0.129  & 0.445  \\
         Random Human Tours & 0.503  & 0.423  & 6.900  & 43.091  & 0.333  & 0.547  \\
         \hline
         PointerNet $p_\text{optimal}(\pi|\mathbf{x})$ &  &  &  &  &  &  \\
         \quad Greedy decoding & 0.540  & 0.457  & 6.644  & 41.596  & 0.347  & 0.569  \\
         \quad Beam search & 0.555  & 0.472  & 6.346  & 40.348  & 0.355  & 0.582  \\
         \quad Best-of-$N$ ($N$=1) & 0.487  & 0.408  & 7.416  & 44.040  & 0.303  & 0.530  \\
         \quad Best-of-$N$ (10) & 0.565  & 0.476  & 6.242  & 39.079  & 0.347  & 0.588  \\
         \quad Best-of-$N$ (100) & 0.578  & 0.488  & 6.087  & 38.923  & 0.346  & 0.595  \\
         \hline
         PointerNet $p_\text{RL}(\pi|\mathbf{x})$ &  &  &  &  &  &  \\
         \quad Greedy decoding & 0.523  & 0.436  & 6.497  & 40.521  & 0.322  & 0.562  \\
         \quad Beam search & 0.523  & 0.437  & 6.487  & 40.498  & 0.322  & 0.563  \\
         \quad Best-of-$N$ ($N$=1) & 0.521  & 0.435  & 6.512  & 40.558  & 0.321  & 0.561  \\
         \quad Best-of-$N$ (10) & 0.537  & 0.449  & 6.351  & 40.097  & 0.325  & 0.570  \\
         \quad Best-of-$N$ (100) & 0.546  & 0.457  & 6.262  & 39.832  & 0.328  & 0.576  \\
         \hline
         PointerNet $p_\text{optimal+RL}(\pi|\mathbf{x})$ &  &  &  &  &  &  \\
         \quad Greedy decoding & 0.540  & 0.457  & 6.374  & 41.183  & 0.353  & 0.573  \\
         \quad Beam search & 0.544  & 0.460  & 6.314  & 40.964  & 0.355  & 0.576  \\
         \quad Best-of-$N$ ($N$=1) & 0.528  & 0.445  & 6.537  & 41.650  & 0.347  & 0.565  \\
         \quad Best-of-$N$ (10) & 0.565  & 0.478  & 6.052  & 38.883  & 0.364  & 0.593  \\
         \quad Best-of-$N$ (100) & 0.576  & \textbf{0.488}  & \textbf{5.930}  & \textbf{38.245}  & \textbf{0.367}  & \textbf{0.600}   \\
         \hline
         PointerNet $p_\text{human}(\pi|\mathbf{x})$ &  &  &  &  &  &  \\
         \quad Greedy decoding & 0.554  & 0.474  & 6.219  & 40.893  & 0.374  & 0.585  \\
         \rowcolor{gray!10}\quad Beam search & 0.575  & 0.493  & 5.948  & 39.370  & 0.388  & 0.600  \\
         \quad Best-of-$N$ ($N$=1) & 0.448  & 0.372  & 7.742  & 45.599  & 0.300  & 0.507  \\
         \quad Best-of-$N$ (10) & 0.551  & 0.466  & 6.302  & 39.054  & 0.357  & 0.584  \\
         \quad Best-of-$N$ (100) & 0.575  & 0.488  & 6.015  & 38.481  & 0.360  & 0.598  \\
         \hline
    \end{tabularx}
    \vspace{1mm} 
    \textit{Note.}  The highlighted gray row indicates the best-performing PointerNet model, which was pretrained directly on human data and decoded using beam search. Bold numbers indicate the best scores among models that were not directly trained on human data, reflecting the strongest similarity between model-generated and human tours for each metric. To quantify overall similarity, we computed an aggregate score by averaging the five evaluation metrics after min–max normalizing each to the interval [0,1]. For consistency, the directions of LD and FD were reversed so that higher values indicated greater alignment with human behavior, matching the other metrics. Consequently, the aggregate score also ranges from 0 to 1, with higher values indicating greater human-likeness. For beam search, the width was set as 100 for a more extensive exploration. EOR=Edge Overlap Ratio, LCS=Longest Common Substring, LD=Levenshtein distance, FD=Fréchet distance, $\tau$=Kendall’s $\tau$ (see Table \ref{tab:overview_of_metrics} for detailed definitions). Standard deviations are omitted from the table because the large number of observations resulted in negligible variability across estimates, with standard deviations close to zero in nearly all cases.
    \label{tab:model_human_comparison}
\end{table}

As shown in Table \ref{tab:model_human_comparison} (gray row), beam-search decoding of tours from the PointerNet policy trained directly on human solutions, $p_\text{human}(\pi|\mathbf{x})$, achieved the strongest overall alignment with human tours, providing an empirical upper bound on model performance. By preferentially selecting high-probability sequences under the learned policy, beam search reveals that the PointerNet concentrates probability mass on human-like tours that can best reflect regularities shared across participants.

For the search-only models, both optimal tours generated by Concorde and near-optimal tours by Elastic Net outperformed the three canonical heuristic approaches and human-tour baseline, achieving levels of human-likeness only slightly below the best PointNet model. These results indicate a strong geometric correspondence between human and optimal tours, mirroring previous findings that humans routinely produce near-optimal tours in TSP tasks \cite{macgregor1996human}. By contrast, all three canonical heuristics previously proposed as models of human TSP performance performed below the baseline of random human tours, demonstrating that they capture insufficient systematic structure in human solutions, even less than a randomly sampled human tour.

Among all models without access to any human solutions, we find that the PointerNet policy $p_\text{optimal+RL}(\pi|\mathbf{x})$, pretrained on optimal solutions and subsequently fine-tuned with RL, produced the most human-like solutions overall. When combined with Best-of-$N$ decoding, this model approached the empirical upper bound defined by the PointerNet trained directly on human tours (see the bold numbers in Table \ref{tab:model_human_comparison}). This finding suggests that supervised pretraining on optimal tours allows the model to acquire rich structural representations of high-quality TSP solutions, while RL fine-tuning further reshapes the policy toward effective tour-length optimization. However, when generalizing to novel TSP instances, the learned policy may still assign non-negligible probability mass to candidate tours that deviate from both optimal and human solutions due to the limited capacity of neural networks. Best-of-$N$ decoding provides a form of test-time search: by sampling multiple candidate tours from the learned policy and selecting among them based on tour-length evaluation, it refines model-generated solutions toward tours that are both shorter and more human-like. More broadly, these results suggest that human-like TSP behavior is best captured not by learning alone, but by the combination of learned structural representations, reinforcement-based improvement, and test-time search under capacity constraints.

\section{Characterizing Geometric Features of Human-like Tours}

Beam search applied to the PointNet policy of $p_\text{human}(\pi|\mathbf{x})$ best captured the overall patterns of human tours, establishing an empirical upper bound on the extent to which human behavior can be reproduced at the instance level. Both search-only optimal approximations and Best-of-$N$ sampling from the other PointerNet policies approached this upper bound to varying degrees. To understand the computational basis of this human-likeness, we next examined the geometric characteristics of the generated tours. Specifically, we seek to understand which geometric features distinguish human-like solutions from alternative computational strategies, and what enables the PointNet policy of $p_\text{optimal+RL}(\pi|\mathbf{x})$ to achieve the closest alignment with human tours when combined with Best-of-$N$ sampling.

To characterize the geometry of human-like tours, we developed eight geometric features spanning two complementary categories. Instance-dependent features capture geometric properties whose values are intrinsically determined by the spatial arrangement of cities in a particular TSP instance: (i) smoothness ($G_{sm}$), the average turning angle along a tour; (ii) turning-angle entropy ($G_{tae}$), the variability of turning angles; (iii) edge-length variance ($G_{elv}$), the variability of edge lengths; and (iv) nearest-neighbor ranking ($G_{nnr}$), the average rank of each selected edge among all feasible outgoing edges. In contrast, instance-independent features capture geometric properties that that consistently distinguish high-quality tours across different TSP instances, irrespective of their specific spatial configurations: (i) convex-hull preservation ($G_{ch}$), the proportion of convex-hull boundary edges preserved by a tour; (ii) optimality gap ($G_{og}$), the relative difference in tour length from the corresponding optimal tour; (iii) 2-opt optimality ($G_{2opt}$), the extent to which a tour satisfies 2-opt local optimality; and (iv) crossing rate ($G_{cr}$), the proportion of crossing edges in a tour (see Appendix \ref{tab:overview_of_geometric_features} for details). Here, the geometric properties of beam-searched tours from $p_\text{human}(\pi|\mathbf{x})$ served as the empirical benchmark for human-likeness. We then compared the corresponding geometric features of tours generated by the two most human-like search algorithms (i.e., Concorde and Elastic Net) and the PointerNet policies decoded using Best-of-$N$ sampling against this benchmark to identify the geometric characteristics underlying their discrepancy or alignment with human-like behavior.

\begin{figure}[h!]
    \centering
    \begin{minipage}{0.14\linewidth}
        \begin{overpic}[width=\linewidth]{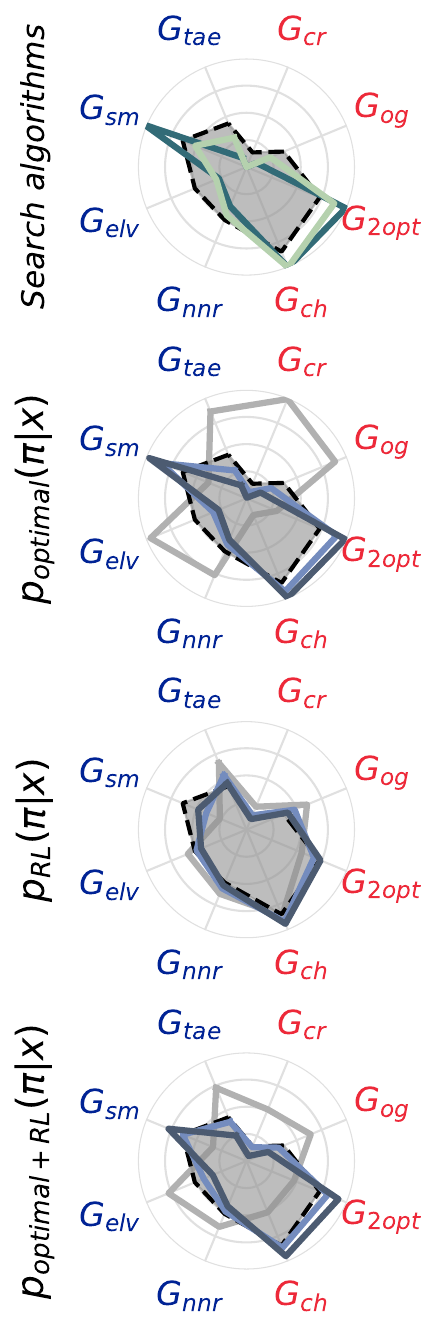}
            \put(3,101){\Large\bfseries a}
        \end{overpic}
    \end{minipage}
    \begin{minipage}{0.852\linewidth}
        \includegraphics[width=\linewidth]{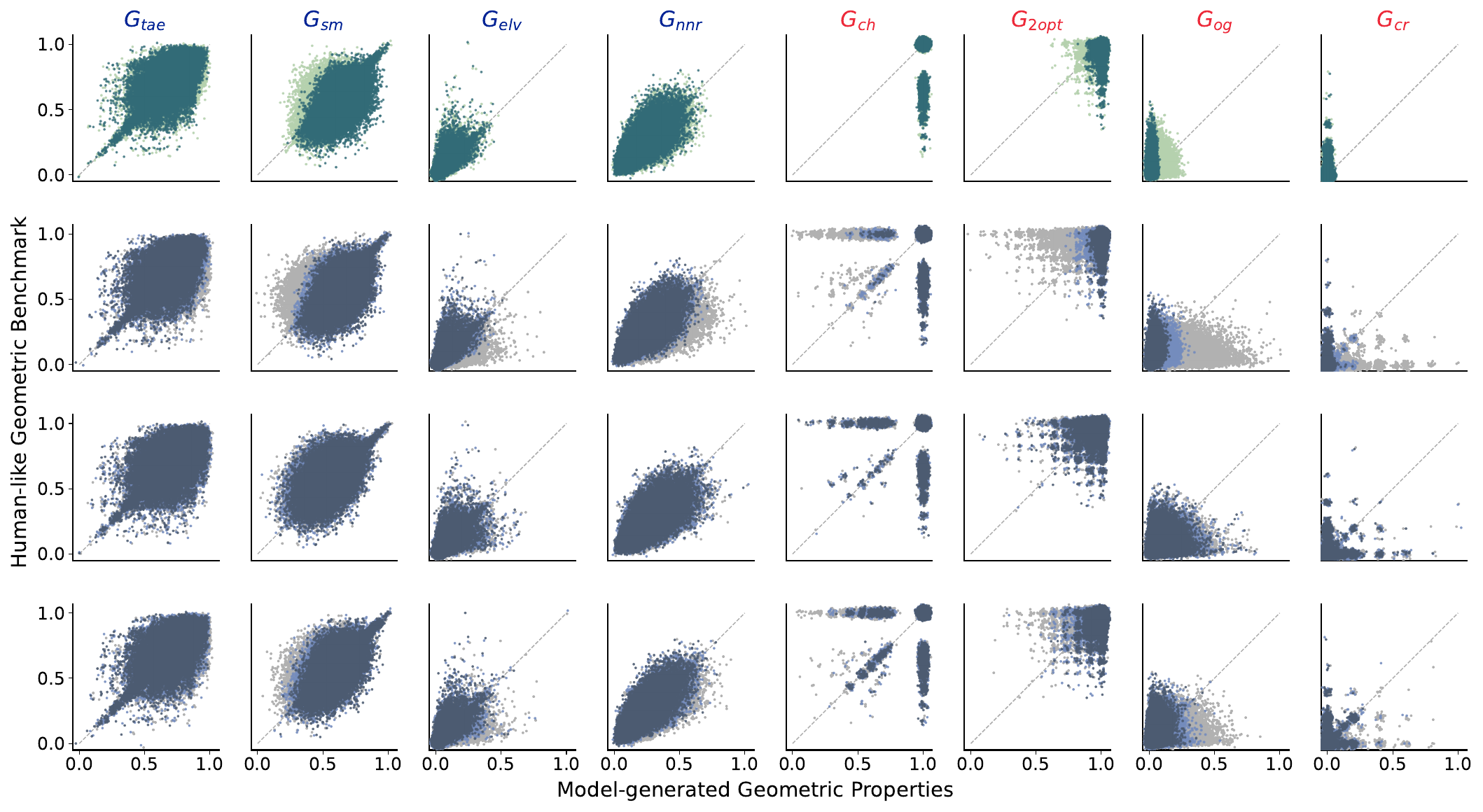}
    \end{minipage}
    \includegraphics[width=0.65\linewidth]{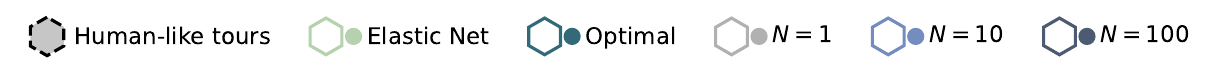}
    
    \par\vspace{3mm}
    \begin{minipage}{0.9\linewidth}
        \begin{overpic}[width=\linewidth]{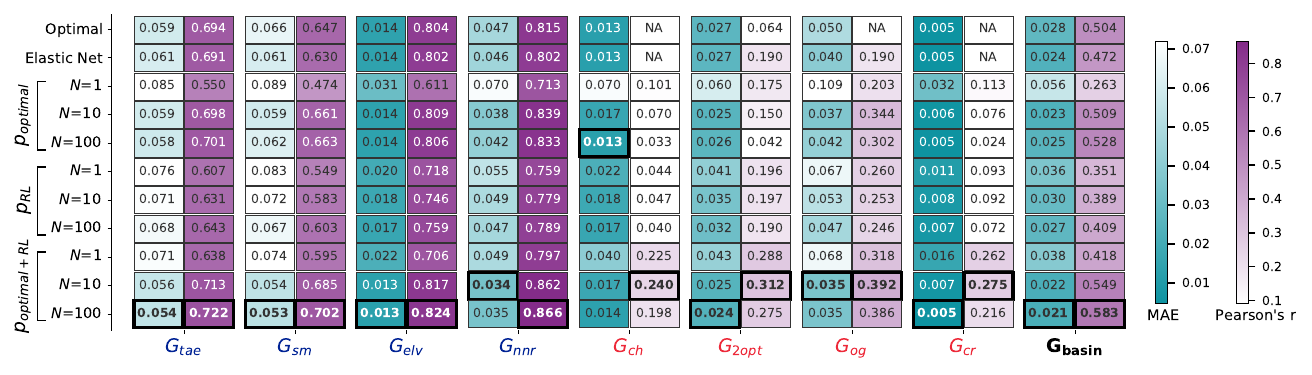}
            \put(-2,27){\Large\bfseries b}
        \end{overpic}
    \end{minipage}
    \caption{
    The comparison between tours decoded from candidate models and the systematic patterns of human tours encoded in the PointerNet policy of $p_\text{human}(\pi|\mathbf{x})$ (the empirical human-like benchmark) along eight geometric properties (blue=instance-dependent properties, red=instance-independent properties): $G_{tae}$=turning-angle entropy, $G_{sm}$=smoothness, $G_{elv}$=edge-length variance, $G_{nnr}$=nearest-neighbor ranking, $G_{ch}$=convex-hull preservation, 
    $G_{2opt}$=2-opt optimality, 
    $G_{og}$=optimality gap,
    $G_{cr}$=crossing rate.
    \textbf{(a)} The radar plots on the left show the average geometric features across instances for each model and decoding strategy included in the comparison, with black dashed lines enclosing the geometric features of human-like tours. The scatter plots on the right further present the instance-level correspondence between candidate models (horizontal axis) and the human-like geometric benchmark (vertical axis). Each dot indicates a TSP instance. Gaussian jitter (standard deviation = 0.01) was added for visualization purposes to reduce overlap among data points.
    \textbf{(b)} Mean absolute errors (MAE) and Pearson's correlation coefficients ($r$) that quantify how each candidate model relates to the human-like benchmark on each geometric feature. The last column applies a geometric basin score that normalized and then aggregated from the eight geometric features. Black boxes with boldface text indicate the model-generated tours that are most closely aligned with the geometry of human-like tours.
    }
    \label{fig:geometric_features_humanlike}
\end{figure}

The extent to which each model aligns with the benchmark is illustrated in Figure \ref{fig:geometric_features_humanlike}a. At the level of TSP instances, both optimal and Elastic Net-derived tours overlap with the geometric region occupied by the human-like tours, while also exhibiting systematic deviations along the extracted geometric features (see Figure \ref{fig:geometric_features_humanlike}a first row). Among the eight geometric features, the most pronounced differences were observed in smoothness, 2-opt local optimality, and convex-hull boundary preservation. On average, compared with tours generated by both the optimal solver and Elastic Net, human-like tours exhibited lower smoothness, lower 2-opt local optimality, and weaker preservation of convex-hull boundaries. These deviations capture characteristic geometric signatures of suboptimality in human-like tours.

In contrast, PointerNet-based approaches exhibited distinct geometries as the size of the Best-of-$N$ sample increased, progressively moving toward the optimal geometric region. For $p_\text{optimal}(\pi|\mathbf{x})$, Best-of-$N$ sampling rapidly drove solutions from a markedly different geometric profile into the optimal region (see Figure \ref{fig:geometric_features_humanlike}a second row). However, tours from $p_\text{RL}(\pi|\mathbf{x})$ aligned well with the human-like benchmark from when $N=1$ and evolved substantially more slowly in subsequent sampling, suggesting that RL produces a strong but comparatively over-concentrated geometric profile that is less responsive to test-time search (see Figure \ref{fig:geometric_features_humanlike}a third row). Sampling from $p_\text{optimal+RL}(\pi|\mathbf{x})$ lay between these two extremes, exhibiting a clear convergence trend toward the optimal region while stabilizing in a near-optimal region adjacent to the human-like benchmark (see Figure \ref{fig:geometric_features_humanlike}a fourth row).

To quantitatively evaluate their alignment with the human-like benchmark in detail, we further checked instance-level correspondence along each geometric feature, as illustrated in Figure \ref{fig:geometric_features_humanlike}a (scatterplots), and summarized model performance using the mean absolute error (MAE) and Pearson's correlation coefficient ($r$) shown in Figure \ref{fig:geometric_features_humanlike}b. Overall, tours generated from both the optimal TSP solver and Elastic Net provided little support for human-like tour geometries on instance-independent features (e.g., $G_{ch}$ and $G_{cr}$ for which both methods consistently predicted perfect convex-hull preservation and complete absence of edge crossings) because human-like tours show systematic suboptimalities in these instances. In contrast, PointerNet-based approaches showed a greater capacity to capture these characteristic suboptimalities in human tours. Best-of-$N$ samples from $p_\text{optimal+RL}(\pi|\mathbf{x})$ yielded the closest alignment with human-like tour geometries, achieving the highest Pearson's correlation across all eight geometric features and the lowest mean absolute error for seven of them. By normalizing and aggregating the eight geometric features into a single geometric basin score, we quantified the overall alignment of each model with the benchmark tours (see the last column in Figure \ref{fig:geometric_features_humanlike}b and Appendix \ref{ap:geo_properties}). Best-of-$N$ samples ($N=100$) from $p_\text{optimal+RL}(\pi|\mathbf{x})$ achieved the strongest human-likeness in reproducing the geometries of human tours, yielding the highest correspondence with the benchmark solutions (Pearson's $r = 0.583$, $p < 0.001$) and the lowest prediction error (MAE = 0.021).

Why does the PointNet policy of $p_\text{optimal+RL}(\pi|\mathbf{x})$ generate more human-like tours under Best-of-$N$ sampling? Test-time compute such as Best-of-$N$ sampling does not modify the weights of the neural network but selects the highest-reward tours from samples drawn from it at test time. Its effectiveness therefore depends largely on the structure of the learned policy distribution: whether the policy assigns sufficient probability mass to human-like candidate tours and how concentrated that probability mass is around its modal solutions. Supervised learning trains the PointerNet by maximizing the conditional likelihood of demonstration tours (i.e., teacher-forcing training), encouraging the policy to model the empirical distribution of optimal tours \cite{ross_reduction_2011}. Consequently, the learned policy assigns substantial probability mass to multiple geometrically optimal-like tours, allowing Best-of-$N$ sampling to progressively converge to the near-optimal region through test-time search. In contrast, RL directly optimizes expected reward (i.e., negative Euclidean distance), progressively concentrating probability mass around a relatively narrow set of high-reward tours, reducing the diversity of high-quality alternatives available for test-time search \cite{liu_policy_2020, ahmed_understanding_2019, sutton_policy_1999}.

Fine-tuning a supervised policy with RL balances these two effects: it preserves broad support over geometrically near-optimal tours while biasing the distribution toward shorter ones. This intermediate policy distribution enables Best-of-$N$ sampling to converge to a near-optimal region that most closely aligns with human behavior. Figure \ref{fig:2d_geometric_space} in Appendix \ref{ap:model_comparisons_evaluation_metrics} provides a conceptual illustration of these learned policy distributions together with two representative test instances showing their effects on individual TSP problems. Notably, the model was not trained on any human tours, yet its decoded tours nevertheless reproduce human-like geometries \cite[c.f.][]{zhu2025language}. This result suggests a computational analogy to human learning: people may acquire broad geometric knowledge about good TSP solutions through exposure to high-quality or approximately optimal examples, and then refine these representations through reinforcement-like feedback from experience.

\section{Discussion}

Many real-world cognitive tasks require selecting high-quality solutions from combinatorially large search spaces under multiple constraints. These tasks share a common computational structure with the TSP: they require efficient navigation through vast spaces of possibilities while satisfying structural constraints, as in route planning; temporal constraints, as in scheduling; or semantic constraints, as in language production. Although search is clearly part of the solution, our results show that learning a strong proposal distribution that guides and accelerates search is equally important for understanding how human-like solutions emerge. More specifically, human-like performance on such problems can emerge from a two-stage learning process: first, acquiring structural knowledge of high-quality solutions through imitation learning from expert or optimal demonstrations, and subsequently adapting this knowledge through RL based on task-specific feedback. Rather than relying on handcrafted heuristics or exhaustive search, this framework enables efficient generation of candidate solutions while preserving sufficient flexibility for test-time search. Similar combinations of imitation learning and RL have proved highly effective in AI research, where supervised pretraining substantially improves the sample efficiency and generalization of subsequent RL \cite{silver_mastering_2016, brown2020language, limozin2026sft}. In this sense, human performance in the TSP may reflect a cognitive analogue of the general-purpose methods emphasized by Richard Sutton's famous ``bitter lesson'' \cite{sutton2019bitter}: effective solutions often arise not from fixed hand-designed heuristics, but from mechanisms that learn from experience and exploit computation through search.

An important question raised by this account is where the demonstrations required for supervised learning come from. Humans are unlikely to directly observe optimal solutions to large combinatorial problems, especially at the scale used in our supervised training. Instead, these internal representations may be acquired gradually through experience with simpler instances whose optimal solutions are easier to discover, allowing useful geometric features to accumulate over development before being transferred to more challenging problems. One promising direction for future research is to develop curriculum-learning or meta-learning algorithms that support stronger generalization in neural networks under limited-data regimes \cite[c.f.][]{mi2026human, lake2023human, griffiths2019doing}.

Our results also highlight the importance of test-time compute as a crucial yet largely overlooked component of cognitive modeling. Rather than treating a neural network as a static model, we show that test-time search is critical for reproducing human behavior. Varying the amount of test-time computation without changing the neural network’s weights provides a principled mechanism for generating human-like variability in solution quality. Of course, more sophisticated test-time search algorithms could be tested in future work. One promising approach is to apply Markov chain Monte Carlo (MCMC) to draw samples from the sequence-level distribution defined by a trained neural network \cite{karan2025reasoning, hu2026simulated}. Autocorrelated samples decoded using MCMC have also been shown to explain classical cognitive phenomena, such as slow errors \cite{zhu2024autocorrelated}. This perspective, which emphasizes the role of test-time computation in modeling intelligent behavior, again parallels recent advances in large language models, where increasing test-time computation, for example through chain-of-thought reasoning or iterative refinement loops, can substantially improve performance without modifying model parameters \cite{snell2024scaling, wei_chain--thought_2023}. Therefore, human-like performance may arise not only from directly reading out learned representations, but also from the amount of computation devoted to deliberation that builds on them.

\section*{Acknowledgments}

H.Y. acknowledges the Chancellor’s International Scholarship from the University of Warwick for support. J.-Q.Z. acknowledges support from the HKU-100 scheme of the University of Hong Kong. Empirical work was performed while M.M. was at South China Normal University, supported by grants from the STI2030‐Major Projects (2021ZD0204200), the Sino‐German Center for Research Promotion (M‐0705), the National Natural Science Foundation of China (32371100), the Research Center for Brain Cognition and Human Development, Guangdong, China (No. 2024B0303390003), and the Striving for the First‐Class, Improving Weak Links and Highlighting Features (SIH) Key Discipline for Psychology in South China Normal University.

\section*{Author Contributions}

H.Y. and J.-Q.Z. conceptualized the theoretical component of this work, while L.H. and M.M. conceptualized the empirical component, conducted the experiments, and collected the human dataset. H.Y. and J.-Q.Z. performed the analyses and computational modeling and wrote the first draft. All authors commented on and contributed to the manuscript.

\newpage
\begin{appendices}

\section{A Large-Scale Experiment of Human Performance in Traveling Salesman Problems} \label{ap:experimental_method_participants}

The AI revolution of the past decade has inspired new approaches to psychological research. The present study adopts one such approach, known as comprehensive exploration (CE). CE seeks to combine the strengths of traditional experimental psychology with the data-driven model development characteristic of AI \cite{huang2026addressing}. On the one hand, CE is theory-oriented: like experimental psychology, it relies on controlled experiments designed to address specific research questions and aims to generate theoretical insight. On the other hand, it is data-guided: like AI research, it emphasizes the construction of large, high-quality benchmark datasets---for example, a single experiment involving 30,000 hours of human data---as the foundation for model development, followed by iterative model refinement until satisfactory performance is achieved. In short, CE is a data-guided, theory-building approach that uses AI tools and methods in pursuit of experimental psychology's central goal: theoretical understanding.

The CE approach builds on earlier work moving in a similar direction \cite{griffiths2015manifesto, watts2017should, agrawal2020scaling, peterson2021using, zhu2025capturing}. Epistemologically, it is designed to achieve both precision, through formal quantitative modeling, and breadth, through broad stimulus coverage and large-scale experimentation. CE has already been successfully applied in several previous studies in cognitive psychology \cite{huang2023quasi, huang2025comprehensive}.

\subsection{Ethics}
The study was approved by the Human Research Ethics Board of South China Normal University and complied with the Declaration of Helsinki. The experiment was conducted as an online game that participants accessed using their personal devices. To obtain informed consent, participants were explicitly informed that the results of the task would contribute to a scientific study. They were also provided with details about the task embedded in the experiment. Participants expressed their willingness to take part in the study by tapping the ``Continue'' button on their personal devices.

\subsection{Online Data Collection Platform}

The current experiment was conducted as an online game using an online data collection platform (\url{https://huang.psy.cuhk.edu.hk/games/}) maintained by L.H. The webpage for data collection was implemented in JavaScript, Vue.js (version 2.6.11), and PHP (version 5.3.3). The platform was embedded within the WeChat app, such that the webpage functioned properly only when accessed through WeChat. This integration was important for facilitating user engagement and participant management.

WeChat is a widely used multipurpose instant messaging and social media application among Chinese users, with over one billion active users. Its broad adoption made it convenient for participants to share the platform through WeChat Moments. More importantly, embedding the platform within WeChat allowed us to use WeChat's identification system, which streamlined participant payment and helped prevent the creation of multiple user IDs. Because the platform relied on automatically detecting WeChat IDs, the webpage had to be accessed exclusively through the WeChat app.

\subsection{Participants}

As described above, the experiment was conducted as an online game that participants accessed using their personal devices, suggesting that most participants were likely active internet users. In addition, the data collection platform was presented in Chinese and embedded within the WeChat app, indicating that participants were likely Chinese-language users. Individuals with color vision deficiencies were explicitly instructed not to participate. Beyond these factors, we are not aware of any other apparent sources of bias in the study population.

A total of 1,232 participants participated in the game. During each week-long session, several hundred active participants received participation-based awards of 25 Chinese Yuan each. Specifically, participants earned one credit for each successfully completed block of 40 trials, with each block taking approximately 5 minutes to complete. Participants were then ranked according to their cumulative credits. At the end of each week-long session, a cutoff was determined based on the cumulative credits of the participant ranked 200th among all participants. Participants whose credits exceeded 75\% of this cutoff, typically corresponding to approximately 300 participants, received a prize of 25 Chinese Yuan. In addition, two participants were randomly selected from among the top 200 and each received a lottery-based award of 500 Chinese Yuan.

\begin{figure}[h!]
    \centering
    \includegraphics[width=\linewidth]{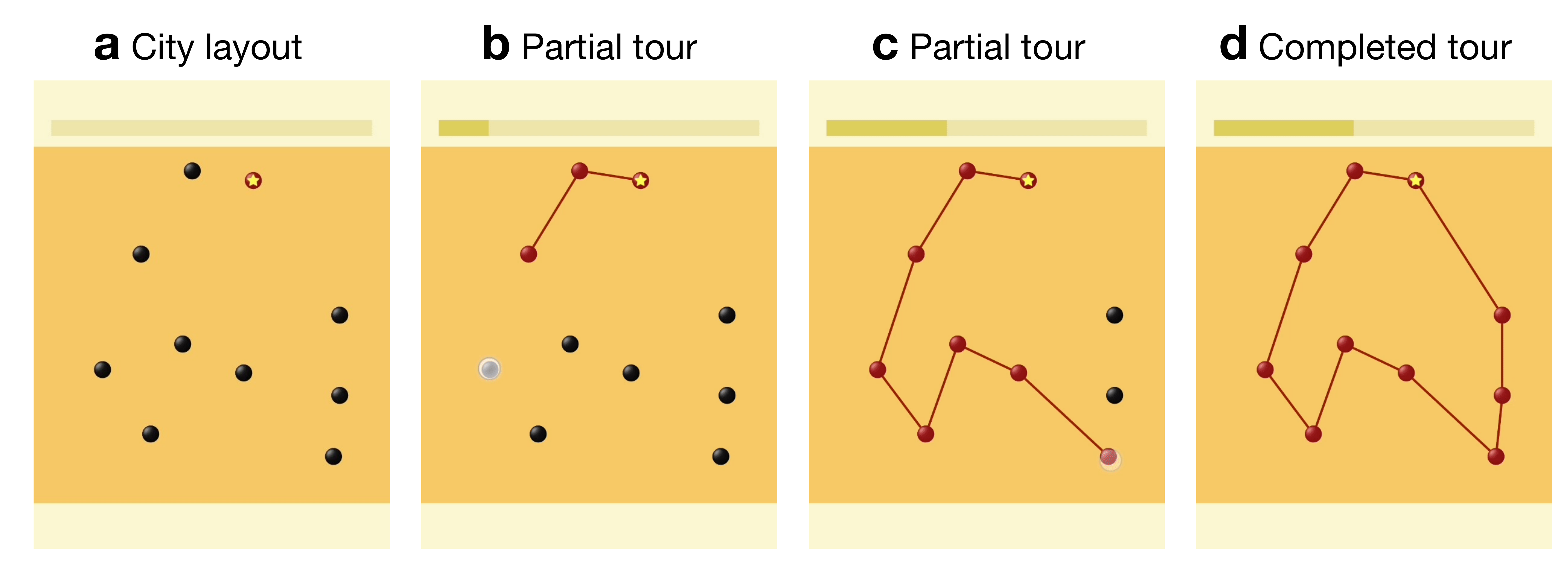}
    \caption{Screenshots of a typical Euclidean TSP trial from the participant's viewpoint. 
    \textbf{(a)} Participants first viewed a two-dimensional city layout, with the required starting city marked by a star. 
    \textbf{(b--c)} Participants constructed a tour sequentially by selecting one city at a time, with selected cities and connecting edges shown in red. 
    \textbf{(d)} The trial ended once all cities had been visited, producing a complete tour.}
    \label{fig:a_typical_trial}
\end{figure}

\subsection{Procedures}

The experiment was administered between May 29, 2023, and September 18, 2023, totaling 16 week-long sessions. Participants were allowed to join the gamified TSP task at any time during this period (see Figure \ref{fig:a_typical_trial} for screenshots of a typical trial). They were instructed to produce short tours. The TSP trials were organized into blocks, with each block consisting of 40 trials: 8 TSP instances from each of the five complexity levels. Complexity levels were defined by the number of cities in each TSP instance: $n = 10, 12, 15, 19,$ and $24$. TSP instances were randomly sampled from the pool of 150,000 instances to satisfy the requirements of each block, and their order within each block was fully randomized.

For each TSP trial, we imposed a time limit equal to twice the number of cities. Thus, for the five complexity levels, the time limits were 20, 24, 30, 38, and 48 seconds, respectively. If a participant failed to complete a tour before the time limit, both the trial and the block containing that trial were recorded as failed. Blocks containing incomplete tours were excluded from the analyses. No feedback about tour length was provided during the experiment.

\subsection{The \texttt{tsp150k} Dataset}

After excluding failed blocks, we obtained the \texttt{tsp150k} dataset, which contains a total of 20,786,680 tours produced by 1,107 participants (404 male and 703 female; mean age = 35.5 years, s.d. = 11.3). The dataset includes 150,000 TSP instances across five complexity levels, with 30,000 instances per level. For each TSP instance, between 92 and 207 human tours were collected, with an average of 139 human tours per instance.

\section{Stylized Facts of Human Performance}
\label{ap:stylized_facts}

\begin{figure}[h!]
    \centering
    \includegraphics[width=\linewidth]{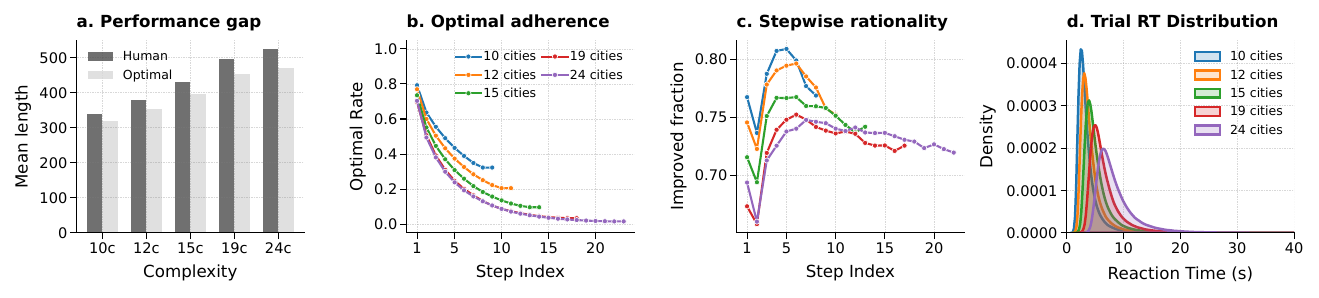}
    \includegraphics[width=\linewidth]{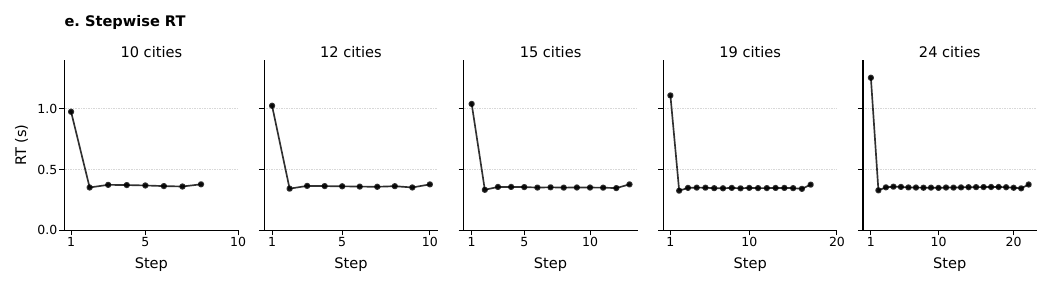}
    \caption{Stylized facts of human performance in TSP. 
    \textbf{(a)} Performance gap between human and optimal tours, measured by Euclidean tour length, across the five complexity levels.
    \textbf{(b)} Proportion of trials in which participants followed the optimal solution up to a given step, shown separately for each complexity level. 
    \textbf{(c)} Stepwise rationality measure: at each step of a trial, conditioned on the completed partial tour, we computed the fraction of available improvement over chance achieved by participants’ current choices.
    \textbf{(d)} Distribution of trial-level response times for each complexity level.
    \textbf{(e)} Mean response time at each step for each complexity level.
    Error bars are omitted because the large number of observations renders sampling uncertainty negligible relative to the reported effect sizes.
    }
    \label{fig:stylized_fact}
\end{figure}

With behavioral data collected at this scale, we can establish several stylized facts about human performance in the Euclidean TSP. Below, we report a series of descriptive analyses that complement our main modeling work.

First, human tours are typically not optimal, but they are near-optimal (see Figure \ref{fig:stylized_fact}a). Across the five complexity levels, the mean ratios of human tour length to the corresponding optimal tour length were 1.062 (10 cities), 1.072 (12 cities), 1.084 (15 cities), 1.098 (19 cities), and 1.110 (24 cities). We further identified, for each TSP instance, the shortest tour produced by any participant. The mean ratios of these best human tour lengths to the corresponding optimal tour lengths were 1.000 (10 cities), 1.000 (12 cities), 1.000 (15 cities), 1.001 (19 cities), and 1.003 (24 cities). Overall, human performance is close to optimal in the TSP: on average, human tours are less than 11\% longer than the optimal tour, and the best human tours are less than 0.3\% longer than the optimal tour.

Second, we calculated the optimality rate of human tours. For a given TSP instance, this rate was computed as the proportion of human tours which followed the optimal tour up to a given time step. For example, if 80\% of human tours selected the same first city as the optimal tour, the optimality rate would be 80\% at time step 1. Subsequently, if 70\% of human tours selected the same first two cities as the optimal tour, the optimality rate would be 70\% at time step 2. As shown in Figure \ref{fig:stylized_fact}b, this rate decreases as more cities are connected in the tour. Moreover, as TSP complexity increases from 10 to 24 cities, the rate of optimal completion decreases approximately to zero.

Third, we examined a local metric of optimality that measures next-city optimality after removing all previously visited cities. That is, we asked whether, ignoring the sunk cost of the current partial tour, the next city chosen by humans would still belong to an optimal tour if the tour were restarted from the current city. Because removing previously visited cities also reduces the number of unvisited cities as participants progress through the task, the random baseline changes accordingly and the task becomes easier over time. To adjust for this decreasing difficulty, we calculated the fraction of improvement over the random baseline for each next-city choice. The random baseline is simply $\frac{1}{\text{number of unvisited cities}}$, assuming that the next city is selected uniformly at random. The improvement fraction is then defined as the proportion of next-city choices that belong to an optimal tour after removing previously visited cities, minus the random baseline, divided by $1$ minus the random baseline. We found that this local optimality rate decreases slightly for the second city, but then rises to a stable level for all subsequent cities (see Figure \ref{fig:stylized_fact}c).


Fourth, on average, participants spent more time on more complex TSP instances. Mean response time (RT) showed an approximately linear relationship with the number of cities in a TSP instance (see Figure \ref{fig:stylized_fact}d): 3.5s (10 cities), 4.3s (12 cities), 5.3s (15 cities), 6.7s (19 cities), and 8.7s (24 cities). This effect may be partially explained by the fact that more complex TSP instances require participants to make more sequential choices, while also increasing the number of available options at each step and expanding the combinatorial space of possible tours factorially with the number of cities.

Fifth, by examining the RT series within each TSP instance, we found that participants spent the most time on the first city, whereas RTs for the remaining cities were substantially shorter and relatively similar across all five complexity levels (see Figure \ref{fig:stylized_fact}e). This pattern suggests that most planning or deliberation may occur at the beginning of the tour, whereas the relatively stable RTs at later steps may primarily reflect the motor implementation of an initially planned route. This RT result also qualitatively aligns with our test-time computation account in the PointerNet model: Best-of-$N$ sampling draws and evaluates $N$ candidate tours from the initial position, which would naturally produce longer deliberation time at the first city.

\section{Computational Models of the Traveling Salesman Problem}
\label{ap:computational_models}

We systematically compare a variety of computational models that generate valid solutions to the Euclidean TSP. Broadly, these models can be classified into (1) exact search methods, which yield optimal solutions; (2) heuristic search methods; (3) neurally inspired dynamic models; and (4) neural-network-based learning and search models.

\subsection{Formalizing the Euclidean Traveling Salesman Problem}

Let a Euclidean TSP instance be denoted by $\mathbf{x} = \{ x_1, \ldots, x_n \}$, where $x_i \in \mathbb{R}^2$ is the location of city $i$ in two-dimensional space, and $n$ is the number of cities. 
We define the Euclidean distance between cities $i$ and $j$ as
\begin{align}    
    d_{ij} = \parallel x_i - x_j \parallel_2,
\end{align}
where $\parallel x_i - x_j \parallel_2 = \sqrt{ (x_i^{(1)} - x_j^{(1)})^2 + (x_i^{(2)} - x_j^{(2)})^2}$. 

A tour $\pi$ can be written as an ordered sequence
\begin{align}    
    \pi = (\pi_1, \ldots, \pi_n),
\end{align}
where $\pi_t$ denotes the index of the city visited at position $t$ in the tour. 
For example, if $\pi=(3,1,4,2)$, then the tour visits city 3 first, city 1 second, city 4 third, and city 2 fourth. Because a TSP tour is cyclic, the choice of starting city is arbitrary. Thus, after fixing one city as the starting point, the set of possible tours contains $(n-1)!$ permutations. Moreover, due to the symmetry of the Euclidean TSP, each tour and its reverse have the same length, leaving $(n-1)!/2$ unique tours. Therefore, for the simplest ($n=10$) and most complex ($n=24$) conditions, the numbers of unique tours to consider are $181,440$ and approximately $1.29\times 10^{22}$, respectively.

Given a tour $\pi$, the Euclidean tour length is 
\begin{align}
    L(\pi;\mathbf{x}) = \parallel x_{\pi_{n}} - x_{\pi_1} \parallel_2 + \sum_{t=1}^{n-1} \parallel x_{\pi_t} - x_{\pi_{t+1}} \parallel_2
\end{align}
The first term $\parallel x_{\pi_{n}} - x_{\pi_1} \parallel_2$ closes the cycle by returning from the last city back to the first. The optimal tour for a TSP instance $\mathbf{x}$ is then defined as the one that minimize the Euclidean tour length:
\begin{align}
    \pi^*(\mathbf{x}) = \arg\min_{\pi \in S_n} L(\pi;\mathbf{x})
\end{align}

\subsection{Concorde (Exact TSP Solver)}

Although solving TSP exactly is computationally demanding in general, the relatively small problem sizes considered in this study (up to 24 cities) make exact optimization computationally feasible. We therefore generated the optimal solution for each TSP instance using Concorde \cite{david2006concorde}, a state-of-the-art exact TSP solver based on a branch-and-cut framework.

Concorde formulates the TSP as an optimization problem over binary edge variables. For each pair of cities $(i,j)$, the variable $\mathbf{1}_{ij} \in \{0,1\}$ indicates whether the edge connecting the two cities is included in the tour. The objective is to minimize the total tour length while ensuring that every city is visited exactly once and that all selected edges form a single Hamiltonian cycle.

Rather than solving this integer optimization problem directly, Concorde first solves a linear programming (LP) relaxation by allowing each edge variable to take any fractional value between 0 and 1, i.e., replacing the binary constraint $\mathbf{1}_{ij}\in\{0,1\}$ with $0\leq \mathbf{1}_{ij}\leq1$. This relaxation can be solved efficiently but may produce intermediate solutions that do not correspond to valid TSP tours. For example, instead of producing a single tour that visits every city exactly once, the LP may produce multiple disconnected subtours, each visiting only a subset of cities. Concorde detects these violations and introduces additional linear constraints, known as \textit{cutting planes}, that explicitly eliminate the invalid solution while preserving every feasible TSP tour. The LP is then solved again with the new constraints. By repeatedly identifying invalid solutions and adding the corresponding cutting planes, the LP relaxation progressively approaches the feasible set of valid TSP tours.

If the LP relaxation still yields multiple competing candidates, Concorde further partitions the search through \textit{branching}. For example, if the LP cannot determine whether a particular edge should belong to the optimal tour, it creates two subproblems: one in which the edge is required to be included ($\mathbf{1}_{ij}=1$) and another in which it is forbidden ($\mathbf{1}_{ij}=0$). Each subproblem is solved independently using the same branch-and-cut procedure. Subproblems whose best achievable objective value is already longer than the best tour found so far are discarded immediately, allowing large portions of the search space to be eliminated without explicit enumeration.

This procedure guarantees global optimality because every feasible TSP tour remains feasible throughout the optimization. Cutting planes eliminate only infeasible or suboptimal regions of the search space without excluding any valid optimal tour, while branching systematically explores every remaining possibility that could still contain the optimum. Subproblems are pruned only when rigorous mathematical bounds prove that they cannot yield a shorter tour than the best solution already found. Consequently, when the algorithm terminates, every candidate tour that could outperform the reported solution has either been explicitly explored or mathematically excluded, certifying that the returned tour is globally optimal. In practice, Concorde has been successfully applied to TSP instances far larger than those considered in this study, including benchmark instances containing up to 85,900 cities.

\subsection{Heuristic Search}

In contrast to the exact solver, classic heuristic algorithms construct approximate solutions by following simple search rules. Although these algorithms do not guarantee global optimality, they are computationally efficient and have long been proposed as models of human TSP performance because they exploit intuitive geometric properties of Euclidean tours \cite{macgregor1996human,van2003convex,macgregor2011human}. 

\textit{Nearest Neighbor.} The Nearest Neighbor heuristic constructs a tour greedily by repeatedly selecting the closest unvisited city to the latest city. Starting from the designated starting city, the algorithm searches among all unvisited cities and extends the partial tour by choosing the city with the shortest Euclidean distance. This process is repeated until every city has been visited, after which the tour returns to the starting city. Because each decision depends only on the current city and its local neighborhood, Nearest Neighbor represents a purely local search strategy that does not consider the global geometry of the remaining tour.

\textit{Convex Hull Cheapest Insertion}. The Convex Hull Cheapest Insertion heuristic exploits the observation that cities on the boundary of the convex hull frequently appear consecutively in optimal Euclidean tours. The algorithm first computes the convex hull of the city set and connects the boundary cities to form an initial partial tour. It then repeatedly inserts one interior city at a time into the existing tour. At each iteration, the algorithm evaluates every possible insertion position and selects the city-edge combination that produces the smallest increase in total tour length. This procedure gradually expands the tour while preserving its global geometric structure.

\textit{Largest Interior Angle}. Largest Interior Angle follows the same initialization procedure as Convex Hull Cheapest Insertion by first constructing the convex hull and then inserting the remaining interior cities sequentially. However, instead of minimizing the increase in tour length, each interior city is inserted into the edge that subtends the largest interior angle with respect to that city. This geometric criterion encourages the tour to maintain smooth boundary-like trajectories while avoiding sharp detours, producing tours that often resemble optimal Euclidean solutions despite relying solely on local geometric information.

\subsection{Elastic Net}

The Elastic Net algorithm \cite{durbin1987analogue} approaches the TSP by optimizing a continuous geometric representation rather than directly constructing a discrete tour. It formulates the problem as an energy minimization task in which the tour emerges from the equilibrium of competing forces (a helpful visualization of Elastic Net can be found at \url{https://ena-tsp.mathieularose.com/}).

The algorithm begins by initializing a closed ring consisting of more points than the number of cities, typically arranged as a small circle near the center of the TSP instance. Each city exerts an attractive force on every point of the ring, encouraging the ring to pass close to the cities, while neighboring points on the ring exert elastic forces on one another that encourage the ring to minimize its length. These competing forces define an energy function that balances proximity to the cities with smoothness of the ring. Using the default parameter settings, the global minimum of this energy approximates the shortest feasible TSP tour.

The positions of the ring points are then optimized using gradient descent to minimize this energy function. As optimization proceeds, the ring progressively expands, deforms, and wraps around the cities while remaining connected. After convergence, each city is associated with its nearest point on the ring, and the order of these points along the ring determines the final TSP tour.

Because the optimization is performed over a continuous geometric object rather than directly over discrete tours, the Elastic Net can be viewed as searching a continuous relaxation of the TSP. However, the energy landscape is non-convex, making gradient descent generally converge to a local optimum rather than the globally shortest tour. Consequently, the Elastic Net produces high-quality approximate solutions without guaranteeing global optimality.

\subsection{Pointer Network}

Modern neural-network methods for solving the TSP aim to learn a stochastic policy $p_\theta(\pi|\mathbf{x})$, where $\theta$ denotes the trainable parameters of the policy network \cite{vinyals2015pointer,bello2016neural}. The chain rule can be used to further factorize the probability of a tour as

\begin{align}
    p_\theta(\pi|\mathbf{x}) = \prod_{t=1}^n p_\theta(\pi_t| \pi_{<t}, \mathbf{x})
\end{align}
where individual softmax modules can be used to represent $p_\theta(\pi_t| \pi_{<t}, \mathbf{x})$ for all $t \leq n$. This is the familiar autoregressive factorization of the joint distribution used in sequence-to-sequence models \cite{sutskever2014sequence} and large language models \cite{radford2018improving}.

The Pointer Network consists of two recurrent neural network modules: an encoder and a decoder \cite{bello2016neural}. In our implementation, both modules are Long Short-Term Memory (LSTM) cells \cite{hochreiter1997long}. At inference, city locations are first transformed by an linear embedding layer into a sequence of $d$-dimensional embeddings, $\{emb_t\}_{t=1}^n$ ($emb_t \in \mathbb{R}^d$). The encoder then reads the embeddings, one at a time and transforms them into a sequence of latent memory states $\{enc_t\}_{t=1}^n$, where $enc_t \in \mathbb{R}^d$. The input to the first decoder step is the first city, from which our participants were required to start their tours. Similar to the encoder, the decoder network maintains its own sequence of latent memory states $\{dec_t\}_{t=1}^n$, where $dec_t \in \mathbb{R}^d$. At each decoding step, a pointing mechanism is used to produce a distribution over the next city to visit in the current tour. Once the next city is selected, it is passed as the input to the next decoder step.

The pointing mechanism masks out previously decoded cities from further generation, thereby ensuring that the generated tour is a permutation of the cities. This mechanism is similar to the attention mechanism proposed in \cite{bahdanau2014neural}. At each decoding step $t$, the attention function takes as input the decoder memory state $q=dec_t$ as the query vector, and a set of encoder memory states $ref=\{enc_1,\ldots, enc_k\}$ as the reference vectors. The attention function then predicts a distribution $A(q, ref)$ over the set of $k$ references. This probability distribution can be understood as the degree to which the model points to each reference $r_i ~(\text{where~} i\leq k)$ after observing the query $q$.

More specifically, the logit for the $i$-th reference vector produced by the attention mechanism is defined as
\begin{align}
    u_i = 
    \begin{cases}
        v^\top \cdot \text{tanh}(W_{ref}\cdot r_i + W_q \cdot q) ,~&\text{if}~ i\neq \pi_j \text{~for all~} j \leq i \\
        -\infty, &\text{~otherwise}
    \end{cases}
\end{align}
Here, the trainable parameters are $W_{ref}, W_q \in \mathbb{R}^{d\times d}$ and $v \in \mathbb{R}^d$. The predicted probability distribution is then given by a softmax over all $k$ logits: 
\begin{align}
    A(q, ref) = \text{softmax}(\mathbf{u}), ~\text{where~} \mathbf{u}=\{u_1,\ldots,u_k\}
\end{align} 
Therefore, the Pointer Network assigns the probability of visiting the next city $\pi_t$ in the tour as follows:
\begin{align}
    p_\theta(\pi_t| \pi_{<t}, \mathbf{x}) = A(q=dec_t, ref=enc_{1:n}),
\end{align}
that is, by applying the attention mechanism to the decoder memory state at step $t$ and all encoder memory states. Setting the logits of cities that have already appeared in the tour to $-\infty$ ensures that the model only points to cities that have not yet been visited, and therefore outputs valid TSP tours.

Finally, we can control the shape of the probability distribution produced by the attention mechanism by introducing temperature scaling:
\begin{align}
    A(q, ref) = \text{softmax}(\mathbf{u}/T), 
\end{align}
where $T$ is the temperature, which is set to $T=1$ during training. When $T>1$, the distribution becomes less peaked, thereby encouraging more exploratory generations of TSP tours.

\subsubsection{PointerNet Architecture and Hyperparameters}

All PointerNet used throughout this work shared the same network architecture (see hyperparameters included in Table \ref{tab:pointer_supervised_hyperparameters}). Each city was represented by its two-dimensional coordinate and projected into a 128-dimensional embedding space. The embedded sequence was processed by a single-layer LSTM encoder with 128 hidden units. A single-layer LSTM decoder then generated the tour autoregressively, starting from the designated first city. At each decoding step, an attention-based pointing mechanism computed logits over all input cities from the decoder hidden state and the encoder states. Previously visited cities were masked to ensure that each city could be selected only once, so that the decoded sequence formed a valid TSP tour.

\begin{table}[h!]
\centering
\caption{Hyperparameters of PointerNet Architecture.}
\begin{tabularx}{\linewidth}{lX}
\hline
Hyperparameter & Value \\
\hline
Embedding dimension & 128 \\
Encoder & Single-layer LSTM \\
Decoder & Single-layer LSTM \\
Hidden dimension & 128 \\
\hline
\end{tabularx}
\label{tab:pointer_supervised_hyperparameters}
\end{table}

\subsubsection{Training Details}

All training procedures described below were conducted using 50\% of the \texttt{tsp150k} dataset (75,000 TSP instances; 15,000 instances from each of the five complexity levels). An additional one-sixth of the dataset (25,000 instances; 5,000 per complexity level) was reserved as a validation set for early stopping and model selection. The remaining one-third of the dataset (50,000 instances; 10,000 per complexity level) was held out as the test set. Unless otherwise stated, all similarity metrics and geometric analyses reported in this study were computed exclusively on this held-out test set.

We trained the stochastic policy of the Pointer Networks, $p_\theta(\pi|\mathbf{x})$, using supervised learning with two types of teacher tours separately: (1) optimal tours, $\pi^*(\mathbf{x})$, and (2) human-generated tours, $\pi_{human}(\mathbf{x})$. Training the Pointer Network follows the standard teacher-forcing procedure used for RNNs, whereby the decoder received the ground-truth city from the previous decoding step as input, and the model was optimized using the cross-entropy loss between the predicted probability distribution and the next city in the optimal or human-generated tour.

Training was performed using the Adam optimizer \cite{kingma2014adam} with a learning rate of $1\times10^{-4}$ and a batch size of 128. The model was trained for 50,000 parameter-update steps. To stabilize optimization, the gradient norm was clipped to a maximum value of 1.0. Validation loss was evaluated every 100 training steps, and the model with the lowest validation loss was retained for subsequent analyses.

To improve generalization, two forms of data augmentation were applied during training. First, for each TSP instance, the order of the input cities was randomly permuted while keeping the designated starting city fixed, and the corresponding target tour was transformed accordingly. This augmentation encourages the model to learn permutation-invariant representations of TSP instances rather than memorizing particular input orderings. Second, because a TSP tour is equivalent when traversed in either direction, the target tour was randomly reversed with probability 0.5, encouraging the model to learn both traversal directions equally.

We also trained a version of the Pointer Network using reinforcement learning (RL), specifically with a policy-gradient method based on the proximal policy optimization (PPO) algorithm \cite{schulman2017proximal}. In this setting, the Pointer Network defines a stochastic policy over tours, and the reward assigned to each sampled tour is the negative Euclidean tour length. Thus, maximizing expected reward corresponds to minimizing the total length of the generated tour.

The actor network was initialized randomly, and the critic network shared the same encoder architecture and consisted of three glimpse-processing blocks followed by a two-layer multilayer perceptron that predicted the expected tour length. At each training iteration, the actor sampled one tour for each TSP instance, and the tour length was computed as the reinforcement signal. The critic estimated the expected tour length of each instance, and the advantage was computed as the difference between the sampled tour length and the critic prediction. Advantages were normalized within each training batch to improve optimization stability. 

The actor was optimized using the clipped PPO surrogate objective with clipping parameter $\epsilon=0.2$. An entropy regularization term with coefficient 0.01 was added to encourage policy exploration and prevent premature collapse of the policy distribution. The critic was trained by minimizing the mean squared error between its predicted tour length and the observed tour length. Training used the Adam optimizer \cite{kingma2014adam} with an initial learning rate of $1\times10^{-4}$ for both the actor and critic. The learning rate decayed by a factor of 0.96 every 100 training iterations using a step scheduler. Each rollout consisted of 128 sampled tours, followed by three PPO optimization epochs using randomly shuffled minibatches of size 32. Gradient norms for both networks were clipped to a maximum value of 1.0. Training was performed for 200,000 rollout iterations, and the actor achieving the lowest average tour length on the validation set was retained for subsequent experiments.

Finally, we trained the PointerNet using a two-stage procedure. In the first stage, the model was pretrained by supervised learning with teacher forcing on optimal tours. In the second stage, the pretrained model was further fine-tuned using PPO, in which the policy was optimized on tours sampled from the model itself. The PPO training procedure followed that described above and was run for 10,000 rollout iterations. The actor achieving the lowest average tour length on the validation set was retained for subsequent experiments. This two-stage procedure is analogous to the training pipelines used in several leading AI systems, such as AlphaGo \cite{silver2017mastering} and GPT-3 \cite{brown2020language}, where supervised pretraining is followed by RL fine-tuning.

\begin{table}[h!]
    \centering
    \caption{Overview of Pointer Network training methods and objectives.}
    \begin{tabularx}{\linewidth}{lXX}
        \hline
        Model & Training Method & Training Objective \\ 
        \hline
        $p_\text{optimal}(\pi|\mathbf{x})$ & Supervised learning & Predict optimal tours \\
        $p_\text{human}(\pi|\mathbf{x})$ & Supervised learning & Predict human-generated tours \\
        $p_\text{RL}(\pi|\mathbf{x})$ & Reinforcement learning  & Maximize the negative Euclidean length of generated tours \\
        $p_\text{optimal+RL}(\pi|\mathbf{x})$ & Supervised learning with reinforcement learning fine-tuning & Predict optimal tours, followed by fine-tuning on maximizing the negative Euclidean tour length \\
        \hline
    \end{tabularx}
    \label{tab:overview_of_pointer_net}
\end{table}

\subsubsection{Decoding Methods}
\label{ap:decoding_methods}
The decoder of the PointerNet generates a tour autoregressively, selecting one city at a time conditioned on the cities that have already been visited. Thus, although the model defines a probability distribution over complete tours, the actual tour produced by the model also depends on the decoding method used during generation. In other words, tour-level behavior is shaped not only by the learned policy $p_\theta(\pi_t|\pi_{<t}, \mathbf{x})$, but also by how this policy is decoded. As a result, different decoding strategies can produce different tours even for the same trained model and the same TSP instance.

We considered several decoding strategies. The simplest is greedy decoding, in which the model selects the city with the highest probability at each decoding step:
\begin{align}
    \pi_t = \arg\max_i p_\theta(\pi_t=i \mid \pi_{<t}, \mathbf{x}).
\end{align}
Greedy decoding is deterministic and efficient, but it may fail to find high-quality tours because locally optimal choices early in the sequence can constrain later decisions.

We also considered stochastic sampling, in which the next city is sampled from the probability distribution produced by the decoder:
\begin{align}
    \pi_t \sim p_\theta(\pi_t \mid \pi_{<t}, \mathbf{x}).
\end{align}
Compared with greedy decoding, stochastic sampling allows the model to generate a more diverse set of tours for the same TSP instance, thereby characterizing the broader distribution implied by the trained policy. We additionally considered temperature-scaled sampling, where the logits produced by the attention mechanism are divided by a temperature parameter $T$ before applying the softmax. Larger values of $T$ produce flatter and more exploratory distributions, whereas smaller values concentrate probability mass on high-probability cities, with $T=0$ corresponding to greedy decoding.

Building on stochastic sampling, we further considered a Best-of-$N$ decoding strategy. Under this strategy, we sample $N$ complete tours (with temperature $T=1$) from the trained PointerNet and select the tour with the shortest Euclidean length:
\begin{align}
    \pi^{\text{Best-of-}N} = \arg\min_{\pi^{(m)} \in \{\pi^{(1)}, \ldots, \pi^{(N)}\}} L(\pi^{(m)}),
\end{align}
where $L(\pi)$ denotes the Euclidean length of tour $\pi$. Best-of-$N$ decoding therefore allows the model to explore multiple candidate tours while returning a single high-quality solution.

Finally, we considered beam search. Rather than committing to one partial tour at each decoding step, beam search maintains the top $B$ partial tours according to their cumulative log probability under the model. These partial tours are expanded autoregressively, and only the best $B=100$ candidates are retained at each step. Beam search can therefore be viewed as a heuristic approximation to the mode of the tour-level stochastic policy $p_\theta(\pi|\mathbf{x})$, as it attempts to identify a high-probability complete tour without exhaustively enumerating all possible tours:
\begin{align}
    \pi^\text{Beam Search}\approx \arg\max_{\pi} p_\theta(\pi |\mathbf{x})
\end{align}

\begin{table}[h!]
    \centering
    \caption{Overview of decoding strategies applied to trained PointerNets.}
    \begin{tabularx}{\linewidth}{lX}
        \hline
        Decoding Method & Description \\ 
        \hline
        Greedy decoding & Selects the city with the highest next-step probability at each decoding step \\
        Best-of-$N$ & Randomly samples $N$ complete tours and selects the one with the shortest Euclidean length \\
        Beam search & Maintains the top $B$ partial tours and provides a heuristic approximation to the highest-probability tour under the tour-level policy \\
        \hline
    \end{tabularx}
    \label{tab:overview_of_decoding_pointernet}
\end{table}

\section{Model Comparisons}
\label{ap:model_comparisons_evaluation_metrics}

\subsection{Evaluation Metrics}
Since a large variety of computational models will be evaluated, we developed a set of tour-level metrics that target different aspects of model performance (see Table \ref{tab:overview_of_metrics} for an overview). These tour-level metrics evaluate the similarity between model-generated tours and human tours. These global metrics capture broader structural properties of the generated solutions, including shared edges, sequential overlap, geometric similarity, and overall ordering, and therefore provide a more direct assessment of whether a model produces human-like tours as complete solutions. Together, these metrics allow us to evaluate global tour-level correspondence between computational models and human behavior.


\begin{table}[h!]
    \centering
    \caption{Overview of evaluation metrics for computational models of human TSP tours.}
    \begin{tabularx}{\linewidth}{lX}
        \hline
        Metric & Description \\ 
        \hline
        Edge overlap ratio ($\uparrow$) 
        & Measures the proportion of shared edges between model-generated and human tours. \\

        Longest common substring ($\uparrow$) 
        & Measures shared local sequential structure between model-generated and human tours. \\

        Levenshtein distance ($\downarrow$) 
        & Measures the number of edits required to transform a model-generated tour into a human tour, including insertions, deletions, and substitutions. \\

        Discrete Fréchet distance ($\downarrow$) 
        & Measures the geometric similarity between model-generated and human tours. \\

        Kendall's $\tau$ ($\uparrow$) 
        & Measures the rank correlation between model-generated and human tours. \\
        \hline
    \end{tabularx}
    \label{tab:overview_of_metrics}
\end{table}

To quantify the overall similarity between a model-generated tour and human solutions, we developed a composite score by aggregating the five pairwise similarity metrics introduced in the main text: Edge Overlap Ratio (EOR), Longest Common Substring (LCS), Levenshtein Distance (LD), Fréchet Distance (FD), and Kendall's $\tau$.

For a given TSP instance, let $\pi_{model}$ denote the model-generated tour and ${\pi_{human}^{(1)},\ldots,\pi_{human}^{(M)}}$ denote the set of tours produced by the $M$ human participants. Each metric was first computed between $\pi_{model}$ and every human tour independently. Because a TSP tour is equivalent when traversed in either direction, each metric was evaluated for both the forward and reversed human tour, and the better matching orientation was retained. The metric values were then averaged across all human participants:

\begin{align}
\frac{1}{M}
\sum_{i=1}^{M}
s_k\left(\pi_{model},\pi_{human}^{(i)}\right),
\end{align}
where $k\in\{\mathrm{EOR},\mathrm{LCS},\mathrm{LD},\mathrm{FD},\tau\}$.

To place the five metrics on a common scale, each metric was normalized to the interval $[0,1]$ (i.e., $s_k \rightarrow \tilde{s}_k$). Metrics that naturally increase with similarity, namely EOR, LCS, and Kendall's $\tau$, were normalized directly. Specifically, EOR and LCS were divided by the number of cities, while Kendall's $\tau$ was linearly transformed from $[-1,1]$ to $[0,1]$. By contrast, the two distance measures, LD and FD, were converted into similarity scores by reversing their direction before normalization to $[0,1]$, so that larger values consistently indicate greater similarity.

Finally, the aggregate score was defined as the arithmetic mean of the five normalized similarity metrics: $\frac{1}{5}\sum_{k}^{}\tilde{s}_k$. The resulting aggregate score thus lies in the interval $[0,1]$, with larger values indicating greater overall similarity between the model-generated tour and the human solutions.

\begin{figure}[t!]
    \centering
    \includegraphics[
        width=\linewidth,
        trim=0cm 3.5cm 13.5cm 0cm,
        clip
    ]{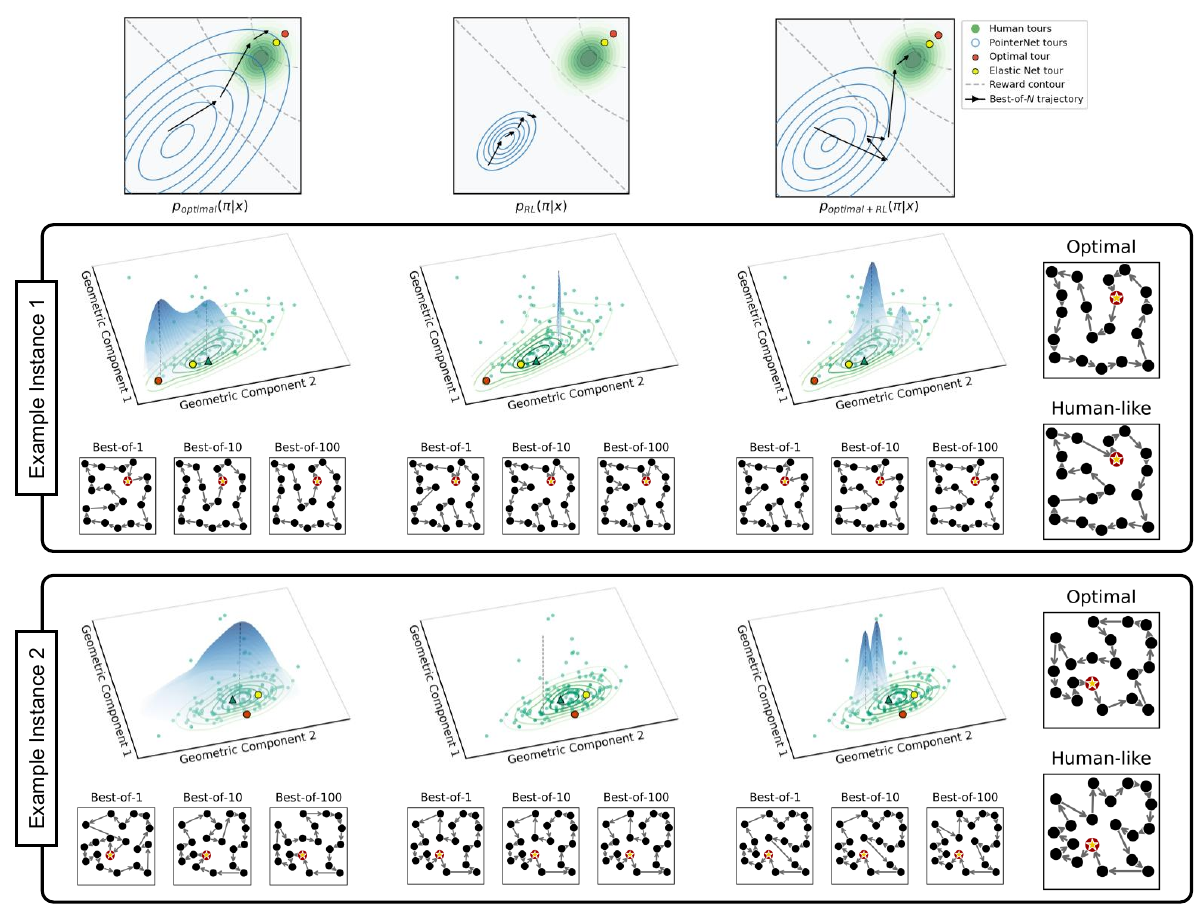}
    \caption{Conceptual illustrations and examples of the geometric distributions learned by different PointerNet policies under Best-of-$N$ sampling. Green dots and contours denote human tours, whereas blue dots and contours denote the tour distributions induced by the PointerNet policy. Each tour was embedded into a vector space defined by eight geometric features and then reduced to two dimensions using t-SNE for visualization. }
    \label{fig:2d_geometric_space}
\end{figure}

\subsection{Geometric Properties of TSP Tours}
\label{ap:geo_properties}

\begin{figure}[t!]
    \centering
    \includegraphics[width=\linewidth]{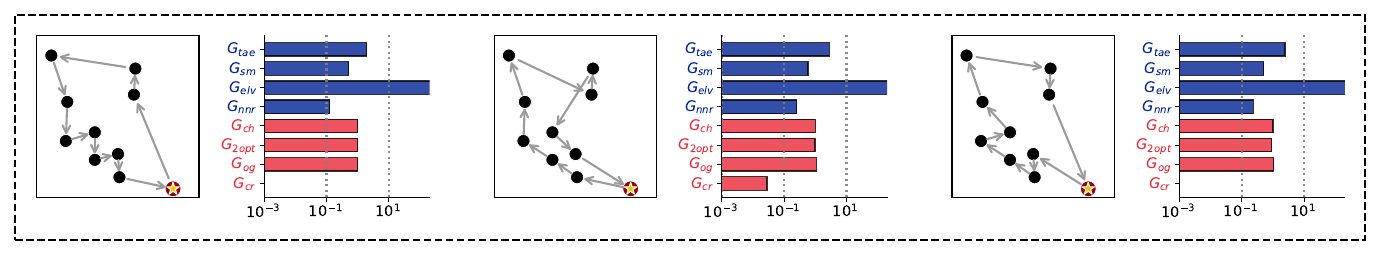}
    \includegraphics[width=\linewidth]{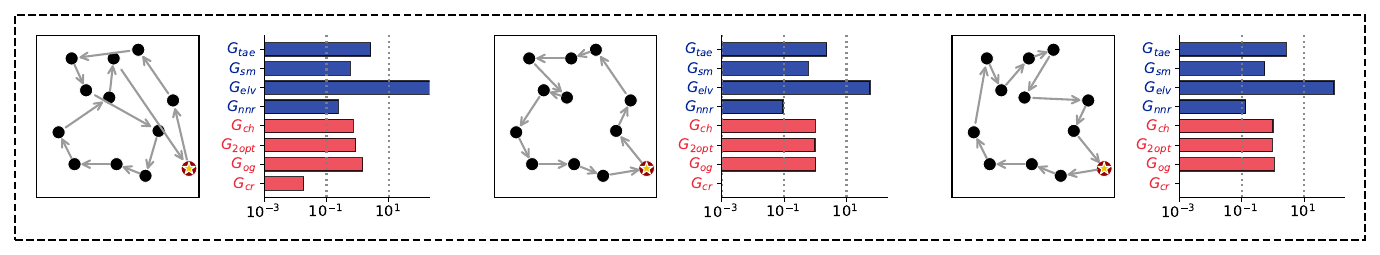}
    \includegraphics[width=\linewidth]{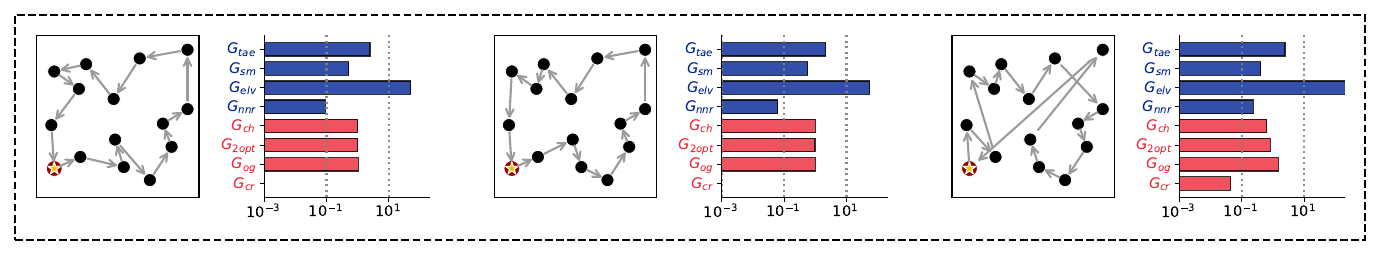}
    \includegraphics[width=\linewidth]{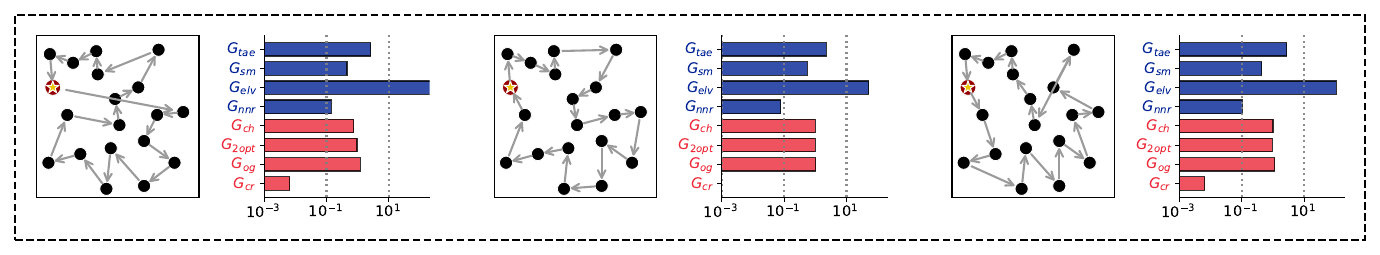}
    \includegraphics[width=\linewidth]{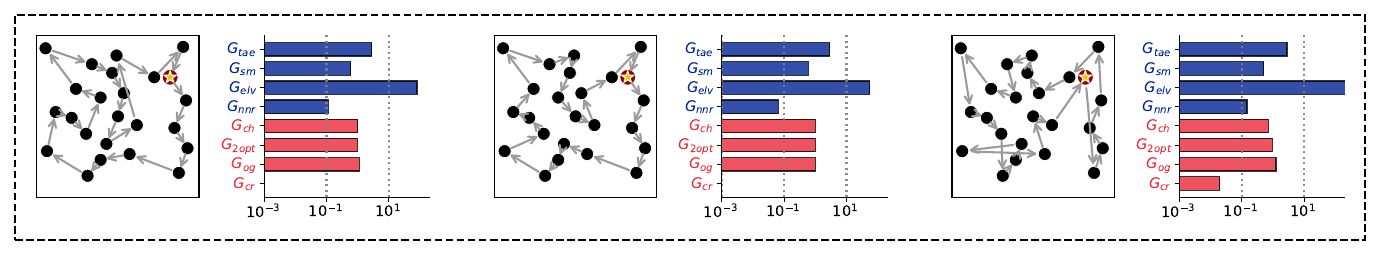}
    \caption{Illustration of the geometric properties of human-generated tours from representative TSP instances. Each row represents a single TSP instance, where three human-generated tours for the instance were randomly selected and visualized together with their corresponding values for the eight geometric features (blue=instance-dependent features, red=instance-independent features). Five TSP instances spanning five levels of problem complexity, ranging from 10 to 24 cities, are shown.}
    \label{fig:human_tours_geometric}
\end{figure}

Why are certain generated tours human-like? To address this question, we developed a series of geometric features that characterize the structure of complete TSP tours. These features are intended to capture qualitative properties often associated with human solutions, such as avoiding unnecessary crossings, connecting nearby cities, producing smooth trajectories, yielding near-optimal distances and preserving convex hull of patterns (see Table \ref{tab:overview_of_geometric_features}). These geometric features can be computed directly from any complete tour (see illustrations in Figure \ref{fig:human_tours_geometric}), allowing us to compare human-like tours and model-generated tours in a shared feature space.

\begin{table}[t!]
\centering
\caption{Overview of the geometric features used to characterize TSP tours.}
\begin{tabularx}{\linewidth}{llX}
\hline
Category & Feature & Description \\
\hline
Instance-dependent  & Turning-angle entropy &
Measures the diversity of turning angles along a tour, reflecting the complexity and regularity of its directional changes. \\

& Smoothness &
Measures the smoothness of a tour by penalizing abrupt changes in travel direction. \\

& Edge-length variance &
Measures the variability of edge lengths, indicating the consistency of step distances throughout the tour. \\

& Nearest-neighbor edge rank &
Measures the extent to which a tour connects each city to its spatially nearest neighbors. \\

Instance-independent & Convex-hull preservation &
Measures the extent to which the tour follows the boundary defined by the convex hull of the point set. \\

& 2-opt local optimality &
Measures the consistency of a tour with local optimality under 2-opt edge-exchange operations. \\

& Optimality gap &
Measures the relative deviation of the tour length from the optimal solution. \\

& Crossing rate &
Measures the frequency with which a tour contains self-intersecting edges. \\
\hline
\end{tabularx}
\label{tab:overview_of_geometric_features}
\end{table}

These features thus provide interpretable summaries of the geometric structure of a tour. They allow us to ask not only whether a model generates tours that are close to human tours according to standard similarity metrics, but also whether the model captures the geometric features that may underlie human TSP behavior. To quantify the overall geometric properties of tours, we developed a Geometric Basin Score by aggregating the eight geometric features. For each geometric feature $G_i$, where
$
G_i\in\{G_{sm},G_{tae},G_{elv},G_{nnr},G_{ch},G_{og},G_{2opt},G_{cr}\},
$
feature values were first normalized across all computational models considered in this study using min-max normalization:

\begin{align}
\tilde{G}_k=\frac{G_k-\min(G_k)}
{\max(G_k)-\min(G_k)},
\end{align}
where the minimum and maximum were computed over all evaluated models. Consequently, every normalized feature lay in the interval ([0,1]).

Because lower values indicate better tours for some geometric features (turning-angle entropy, edge-length variance, nearest-neighbor ranking, optimality gap, and crossing rate), whereas higher values indicate better tours for others (smoothness, convex-hull preservation, and 2-opt optimality), the former were converted into similarity scores by reversing their direction. That is, 
$\hat{G}_k=1-\tilde{G}_k, ~~\text{for}~~
k\in
\{G_{tae},G_{elv},G_{nnr},G_{og},G_{cr}\}
$, whereas $\hat{G}_k=\tilde{G}_k,  ~~\text{for}~~k\in
\{G_{sm},G_{ch},G_{2opt}\}.$

Finally, the Geometric Basin Score was computed as the arithmetic mean of the eight normalized geometric features:
$\frac{1}{8}
\sum_{k}^{}
\hat{G}_k.$
The resulting Geometric Basin Score lies in the interval ([0,1]), with larger values indicating that a tour is located closer to the human-like geometric basin.




\end{appendices}


\bibliography{sn-bibliography}

\end{document}